\documentclass[12pt]{article}
\usepackage{amsmath}
\usepackage{graphicx,psfrag,epsf,color,subfigure}
\usepackage{enumerate}
\usepackage{natbib}
\setcitestyle{numbers}
\setcitestyle{square}

\usepackage[unicode=true,pdfusetitle,
   bookmarks=true, 
   bookmarksnumbered=false, 
   bookmarksopen=false,
   breaklinks=false, 
   pdfborder={0 0 1}, 
   backref=false, 
   colorlinks=false]{hyperref}
 
\usepackage{url} 
\usepackage{amssymb}
\usepackage{multirow}
\usepackage{array}
\usepackage{bm}
\usepackage{booktabs}
\usepackage{float}
\usepackage[font=small,labelfont=bf]{caption}
\setcitestyle{citesep={;}}
\usepackage[total={6.4in,8.6in}]{geometry}
\usepackage{rotating}

\usepackage{subfigure}

\newcommand{\Mean}{{\mbox{E}}}

\newcommand{\prob}{{\mbox{pr}}}
\newcommand\independent{\protect\mathpalette{\protect\independenT}{\perp}}
\def\independenT#1#2{\mathrel{\rlap{$#1#2$}\mkern2mu{#1#2}}}

\newtheorem{theorem}{Theorem}[section]

\DeclareMathOperator*{\argmax}{arg\,max}
\usepackage{authblk}

\begin{document}

\def\spacingset#1{\renewcommand{\baselinestretch}%
{#1}\small\normalsize} \spacingset{1}
%
%
\title{\bf CAPITAL: Optimal Subgroup Identification via Constrained Policy Tree Search}
\author[1]{Hengrui Cai\thanks{
Correspondence to: Hengrui Cai $<$\textcolor{blue}{hengrc1@uci.edu}$>$.} } 
\author[2]{Wenbin Lu}
\author[3]{Rachel Marceau West}
\author[3]{\\Devan V. Mehrotra}
\author[3]{Lingkang Huang}
\affil[1]{Department of Statistics, University of California Irvine}
\affil[2]{Department of Statistics, North Carolina State University}
\affil[3]{Biostatistics and Research Decision Sciences, Merck \& Co., Inc.} 
 \date{}
 \maketitle  
 
\baselineskip=21pt

%


\begin{abstract}
Personalized medicine, a paradigm of medicine tailored to a patient's characteristics, is an increasingly attractive field in health care. An important goal of personalized medicine is to identify a subgroup of patients, based on baseline covariates, that benefits more from the targeted treatment than other comparative treatments. Most of the current subgroup identification methods only focus on obtaining a subgroup with an enhanced treatment effect without paying attention to subgroup size. Yet, a clinically meaningful subgroup learning approach should identify the maximum number of patients who can benefit from the better treatment. In this paper, we present an optimal subgroup selection rule (SSR) that maximizes the number of selected patients, and in the meantime, achieves the pre-specified clinically meaningful mean outcome, such as the average treatment effect. We derive two equivalent theoretical forms of the optimal SSR based on the contrast function that describes the treatment-covariates interaction in the outcome. We further propose a ConstrAined PolIcy Tree seArch aLgorithm (CAPITAL) to find the optimal SSR within the interpretable decision tree class. The proposed method is flexible to handle multiple constraints that penalize the inclusion of patients with negative treatment effects, and to address time to event data using the restricted mean survival time as the clinically interesting mean outcome. Extensive simulations, comparison studies, and real data applications are conducted to demonstrate the validity and utility of our method.
\end{abstract}

\noindent%
\vfill

\section{Introduction}
 
\textcolor{black}{Personalized medicine, a paradigm of medicine tailored to a patient's characteristics, is an increasingly attractive field in health care \citep{kosorok2019precision}. Its ultimate goal is to optimize an outcome of interest by assigning the right treatments to the right patients.  
The success of personalized medicine relies on using baseline covariates to identify a subgroup of patients that benefit more from the targeted treatment than other comparative treatments \citep{loh2019subgroup}. 
The resulting identification strategy is referred to as a subgroup selection rule (SSR). If used properly, subgroup analysis can lead to more well-informed clinical decisions and improved demonstration of the efficacy of a treatment.}

\textcolor{black}{Various data-driven methods for subgroup identification \citep{song2004evaluating,su2009subgroup,foster2011subgroup,cai2011analysis,sivaganesan2011bayesian,imai2013estimating,loh2015regression,fu2016estimating} have been developed during the past decade.  
Song and Pepe (2004) \citep{song2004evaluating} considered using a selection impact curve to evaluate treatment policies for a binary outcome based on a single baseline covariate.  Foster et al. (2011) \citep{foster2011subgroup} and Cai et al. (2011) \citep{cai2011analysis} developed notable methods for detection of subgroups with enhanced treatment effects based on multiple baseline covariates. Foster et al.'s method \citep{foster2011subgroup}, virtual twins (VT), is a two-stage method, first predicting the counterfactual outcome for each individual under both the test and control treatments, and then using tree-based methods to infer the responding subgroups. Cai et al.'s method \citep{cai2011analysis} proposed instead to use a parametric scoring system to rank treatment effects, using this ranking to identify patients who benefit more from the new treatment. 
A useful tutorial and literature review of some commonly used subgroup identification methods is provided in Lipkovich et al. (2017) \citep{lipkovich2017tutorial}. }

\textcolor{black}{
Recently, VanderWeele et al. (2019) \citep{vanderweele2019selecting} considered selecting the optimal subgroup under different constraints, including constrained resources, unconstrained resources, and in the presence of side effects and costs, aiming to maximize effect heterogeneity. This idea of constrained subgroup optimization is becoming increasingly of interest, with many newly proposed methods focusing on performing subgroup selection while simultaneously minimizing patient risk or cost. One such approach was proposed by Wang et al. (2018) \citep{wang2018learning}, who uses outcome weighted learning to generate an individualized optimal decision rule that maximizes the clinical benefit for patients while controlling the risk of adverse events. Guan et al. (2020) \citep{guan2020bayesian} proposed to estimate the optimal dynamic treatment regime under a constraint on the cost function by leveraging nonparametric Bayesian dynamics modeling with policy search algorithms. Zhou et al. (2021) \citep{zhou2021restricted} extended the constrained optimal treatment regime approach to competing risk data, using a penalized value search method to handle the trade-off between the primary event of interest and the time to a severe treatment side effect. Most recently, Doubleday et al. (2022) \citep{doubleday2022risk} proposed two methods to identify risk-controlled individualized treatment rules that maximize benefit while controlling risk at a pre-specified threshold. While these methods make important contributions to finding optimal subgroups under constraints and while balancing risks, they all focus on optimizing the mean outcome of interest without considering the size of the subgroup. In such, they usually yield a smaller, and thus less satisfactory group of selected patients. }

\textcolor{black}{
Identifying the largest possible subgroup of patients that benefit from a given treatment at or above some clinically meaningful threshold can be critical both for the success of a new treatment and more importantly for the patients who may rely on a treatment for their health and survival. When too small of a subgroup is selected, the erroneously unselected patients may suffer from suboptimal treatments. For a test treatment, this reduced subgroup size can further lead to problems with regulatory approvals and may even halt compound development and availability. Post-approval accessibility can also be hindered by a lackluster subgroup size, especially in countries with all-or-nothing reimbursement markets where the seemingly small proportion of benefiting patients leads to reduced reimbursements that may not be financially sustainable for continued treatment manufacturing. Though these points are important, the crucial point remains that a subgroup learning approach that selects as many patients with evidence of clinically meaningful benefit from treatment as possible is desirable to ensure that more patients can receive the treatment that is best for them.  Further, most of the existing optimization approaches with constraints use complex decision rules. It is hard to search within an interpretable class of decision rules using these methods since the loss functions in both outcome weighted learning and value search methods are defined based on the whole sample. Forming an interpretable decision rule helps regulators, doctors, and patients make sense of and ensure proper prescriptions.  }

 \textcolor{black}{
 In this paper, we develop a ConstrAined PolIcy Tree seArch aLgorithm (CAPITAL) to optimize subgroup size while maintaining a pre-specified clinical threshold within the selected subgroup(s). Our contributions can be summarized as follows. First, we derive two equivalent theoretical forms of the optimal SSR based on the contrast function that describes the treatment-covariates interaction in the outcome. Second, we transform the loss function of the constrained optimization into individual rewards defined at the patient level. This enables us to identify the patients with a larger mean outcome and develop a decision tree to generate an interpretable subgroup using the policy tree algorithm proposed by Athey and Wager (2017) \citep{athey2021policy}. Third, we extend our proposed method to the framework with multiple constraints, e.g., penalizing the inclusion of patients with negative treatment results, and to time to event data, using the restricted mean survival time as the clinically interesting mean outcome. Extensive simulations, comparison studies, and real data applications are conducted to demonstrate the validity and utility of our method.  The source code, implemented in the \textsf{R} language, is publicly available at our repository at \url{https://github.com/HengruiCai/CAPITAL}. }

The rest of this paper is organized as follows. We first formulate our problem in Section \ref{sec:prob}. In Section \ref{sec:method}, we establish the theoretical optimal SSR that achieves our objective, and then propose CAPITAL to solve the optimal SSR. We extend our work to multiple constraints and survival data in Section \ref{sec:extens}. Simulation and comparison studies are conducted to evaluate our methods in Section \ref{sec:simu}, followed by the real data analysis in Section \ref{sec:real}. In Section \ref{sec:con}, we discuss and conclude our paper. All the technical proofs and additional simulation results are provided in the appendix.

 \section{Problem Formulation}\label{sec:prob}

Let $X=[X^{(1)},\cdots,X^{(r)}]^\top$ denote a $r$-dimensional vector containing individual's baseline covariates with the support $\mathbb{X} \in \mathbb{R}^r$, and $A \in\{0,1\}$ denote the binary treatment an individual receives. After a treatment $A$ is assigned, we observe the outcome of interest $Y $ with support $\mathbb{Y} \in \mathbb{R}$. Let $Y^*(0)$ and $Y^*(1)$ denote the potential outcomes that would be observed after an individual receives treatment 0 or 1, respectively. Define the propensity score function as the conditional probability of receiving treatment 1 given baseline covariates $x$, denoted as $\pi(x)=\prob(A=1|X=x)$. Denote $n$ as the sample size. The sample consists of observations $\{O_i=(X_{i},A_{i},Y_ {i}), i = 1, \dots , n\}$ independent and identically distributed (I.I.D.) across $i$.

 As standard in the causal inference literature \citep{rubin1978bayesian}, we make the following assumptions: 

\noindent \textbf{(A1)}. Stable Unit Treatment Value Assumption (SUTVA): 
$Y= A Y^*(1)  + (1-A)Y^*(0).$

\noindent \textbf{(A2)}. Ignorability: 
$\{Y^*(0),Y^*(1) \} \independent A\mid X.$
 
\noindent \textbf{(A3)}. Positivity: $0<\pi(x)<1$ for all $x \in \mathbb{X}$.

Based on assumptions (A1) and (A2), we define the contrast function as 
\begin{equation*} 
C(X)\equiv \Mean\{Y^*(1)| X \}-\Mean\{Y^*(0)| X \} =\Mean(Y|A=1, X )-\Mean(Y|A=0, X ),
\end{equation*}
that describes the treatment-covariates interaction in the outcome. Under assumptions (A1) to (A3), the contrast function $C(X)$ is estimable from the observed data. Define the subgroup selection rule (SSR) as $D(X)$ that assigns the patient with baseline covariates $X$ to the subgroup ($D(X)=1$) or not ($D(X)=0$). Denote the class of the SSR as $\Pi$. The goal is to find an optimal SSR that maximizes the size of the subgroup and also maintains a desired mean outcome such as the average treatment effect ($\delta$), with a theoretical version as follows,
\begin{eqnarray}\label{theo_objective}
	 \max_{D\in \Pi}\quad   \prob \{D(X)=1\}, \quad \quad \text{s.t.} \quad \Mean\{Y^*(1)|D(X)=1\}-\Mean\{Y^*(0)|D(X)=1\} \ge \delta>0,
\end{eqnarray} 
where $\delta$ is a pre-specified threshold of clinically meaningful average treatment effect. Based on assumptions (A1) and (A2), the constraint in \eqref{theo_objective} can be represented by
\begin{eqnarray*} 
&&\Mean\{Y^*(1)|D(X)=1\}-\Mean\{Y^*(0)|D(X)=1\}\\
=&&\Mean\{Y|A=1,D(X)=1\}-\Mean\{Y|A=0,D(X)=1\}=\Mean\{C(X)|D(X)=1\}\ge \delta>0.
\end{eqnarray*}
 \textcolor{black}{Combining this with \eqref{theo_objective}, we can obtain the empirical version of the optimization objective: 
\begin{eqnarray}\label{objective}
	 \max_{D\in \Pi}\quad  \sum_{i=1}^n \mathbb{I}\{D(X_i)=1\}, \quad\quad\text{s.t.} \quad  \sum_{i=1}^n \widehat{C}(X_i) \mathbb{I}\{D(X_i)=1\} \ge n \delta>0,
\end{eqnarray}
where $\mathbb{I}(\cdot)$ is an indicator function, and $\widehat{C}(\cdot)$ is some estimator of the contrast function $C(\cdot)$.
}

  \section{Method} \label{sec:method}
 In this section, we first establish the theoretical optimal SSR that achieves our objective, and then propose CAPITAL to solve the optimal SSR.
 
  \subsection{Theoretical Optimal SSR}\label{sec:theo}
\begin{figure} [!t] 
	\centering
	\includegraphics[width=0.45\textwidth]{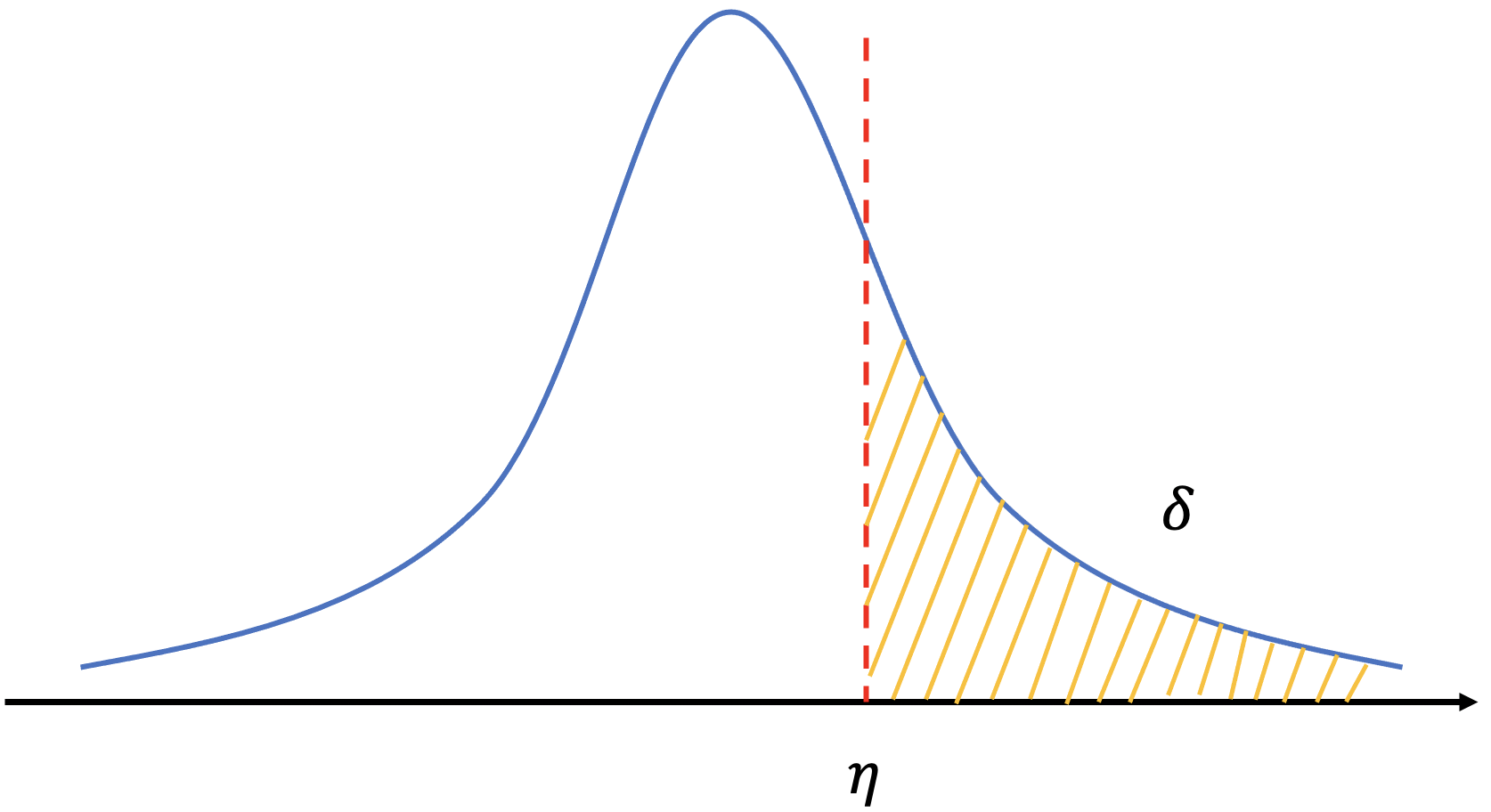}
 	\caption{Illustration of the density function of the contrast function $C(X)$ with a cut point $\eta$ for the pre-specified threshold $\delta$.} \label{fig:theo_ssr}
\end{figure} 
We first derive the theoretical optimal SSR that solves \textcolor{black}{the theoretical objective in \eqref{theo_objective}. }
Given the pre-specified threshold $\delta$, we denote a cut point $\eta$ associated with the contrast function $C(X)$ such that the expectation of the contrast function $C(X)$ larger than $\eta$ achieves $\delta$, i.e.,
\begin{equation}\label{eqn:eta}
\Mean\{C(X)|C(X)\geq\eta\}= \delta.
\end{equation}
By introducing $\eta$, when we are maximizing the subgroup size, the treatment effect of each patient in the subgroup is ensured to meet the minimum acceptable beneficial effect size. We illustrate the density function of the contrast function $C(X)$ with a cut point $\eta$ for the pre-specified threshold $\delta$ in Figure \ref{fig:theo_ssr}. The yellow area in Figure \ref{fig:theo_ssr} contains the patients whose contrast functions are larger than $\eta$ and thus satisfy \eqref{eqn:eta}. 
Intuitively, the theoretical optimal SSR should choose the patients whose contrast functions fall into the yellow area in Figure \ref{fig:theo_ssr}, i.e., those whose treatment effects are larger than $\eta$, to maximize the size of the subgroup. Without loss of generality, we consider the class of the theoretical SSRs as 
\begin{equation*}
\Pi\equiv \left[ \mathbb{I}\{C(X)\geq t\}: t\in \mathbb{R}\right].
\end{equation*}
Here, for a given $t$, the SSR $\mathbb{I}\{C(X)\geq t\}$ selects a patient into the subgroup if his / her contrast function is larger than $t$.
 The following theorem gives the theoretical optimal SSR.

\begin{theorem}{(Theoretical Optimal SSR)}\label{thm1}
Assuming (A1) and (A2), the optimal subgroup selection rule is
\begin{equation}\label{eqn:ossr1}
D^{opt}(x)\equiv\mathbb{I}\{C(x)\geq\eta\}, \forall x \in \mathbb{X}.
\end{equation}
Equivalently, the optimal subgroup selection rule is
\begin{equation}\label{eqn:ossr2}
D^{opt}(x)\equiv\mathbb{I}\left(\Mean_{Z\in \mathbb{X}}[ C(Z) \mathbb{I}\{C(Z)\geq C(x)\}] \geq \delta \right), \forall x \in \mathbb{X} .
\end{equation}
\end{theorem}

The proof of Theorem \ref{thm1} consists of two parts. First, we show the optimal SSR is $   \mathbb{I}\{C(x)\geq\eta\}, \forall x \in \mathbb{X}$, where $\eta$ satisfies \eqref{eqn:eta}, within the class $\Pi$. Second, we derive the equivalence between \eqref{eqn:ossr1} and \eqref{eqn:ossr2}.  See the detailed proof of Theorem \ref{thm1} provided in the appendix.  From Theorem \ref{thm1} and the definition of the cut point $\eta$, the optimal SSR can be found based on the density of the contrast function. Since the density function is usually unknown to us in reality, we use the estimated contrast function for each patient, i.e., the individual treatment effect,  to approximate the density function. A constrained policy tree search algorithm is provided to solve the optimal SSR in the next section.

 \subsection{Constrained Policy Tree Search Algorithm}\label{sec:algo}

 In this section, we formally present CAPITAL. First, we transform the constrained optimization in \eqref{objective} into individual rewards defined at the patient level. This enables us to identify patients more likely to benefit from treatment. Then, we develop a decision tree to partition these patients into the subgroups based on the policy tree algorithm proposed by Athey and Wager (2017) \citep{athey2021policy}. In this paper, we focus on the SSR in the class of finite-depth decision trees. Specifically, for any $L\geq 1$, a depth-$L$ decision tree $DT_L$ is specified via a splitting variable $X^{(j)} \in \{X^{(1)},\cdots,X^{(r)}\}$, a threshold $\Delta_L \in \mathbb{R}$, and two depth-$(L - 1)$ decision trees $DT_{L-1, c_1}$, and $DT_{L-1, c_2}$, such that $DT_L(x) = DT_{L-1, c_1}(x)$ if $x^{(j)} \leq \Delta_L$, and $DT(x) = DT_{L-1, c_2}(x)$ otherwise. Denote the class of decision trees as $\Pi_{DT}$.

 \begin{figure}[!t]  
	\centering
	\includegraphics[width=0.5\textwidth]{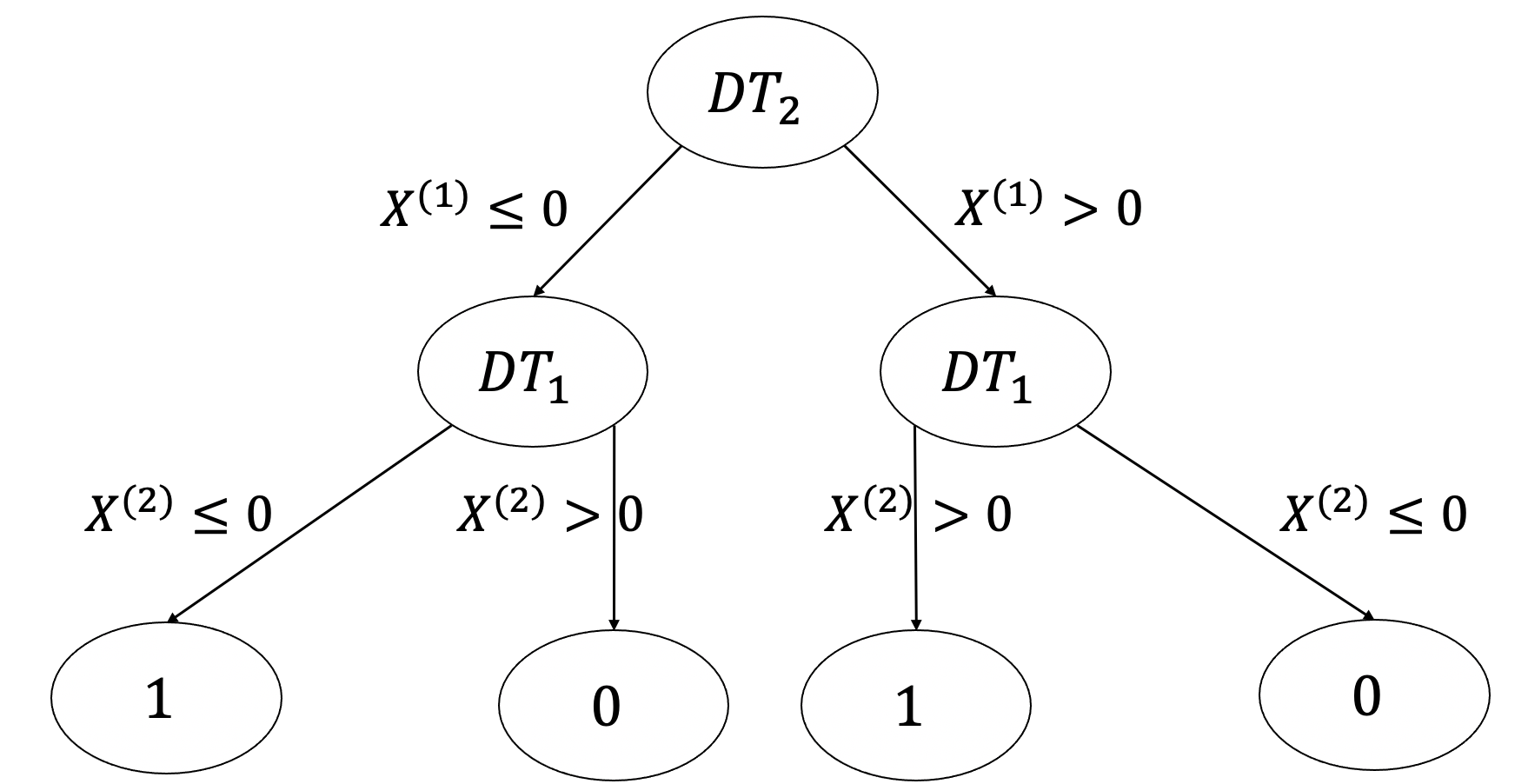}
 	\caption{Illustration of a simple $L=2$ decision tree with splitting variables $X^{(1)}$  and $X^{(2)}$.} \label{fig:dt}
\end{figure}  

We illustrate a simple $L=2$ decision tree with splitting variables $X^{(1)}$  and $X^{(2)}$ in Figure \ref{fig:dt}. This decision tree \textcolor{black}{has mathematical form} $\mathbb{I}\{X^{(1)}X^{(2)}>0\}$. Define $r_i =  C(X_i) -\delta$ as the difference between the contrast function and the desired average treatment effect $\delta$. Under (A1)-(A3), we can estimate the contrast function as $\widehat{C}(\cdot)$, using the random forest method and out-of-bag prediction (see e.g., Lu et al., 2018 \citep{lu2018estimating}). 
 Define $\widehat{r}_i  = \widehat{C}(X_i) -\delta$. It is immediate that a patient with larger $\widehat{r}_i$ is more likely to be selected into the subgroup based on Figure \ref{fig:theo_ssr}. We sort the estimates $\widehat{r}_i$ as 
$\widehat{r}_{(1)}\geq\widehat{r}_{(2)}\geq\cdots\geq\widehat{r}_{(n)}.$ 
This sequence gives an approximation of the density of $C(X) -\delta$. We further define the cumulative mean based on the above sequence $\{\widehat{r}_{(i)}\}$ as
\begin{eqnarray*}
\widehat{R}_{(i/n)} ={1\over i}\sum_{j=1}^i \widehat{r}_{(j)}.
\end{eqnarray*}
With sufficiently large sample size, $\widehat{R}_{(i/n)}$ converges to the average treatment effect minus the desired effect $\delta$, within the selected patients whose contrast function is larger than the upper $i/n$ quantile of the density of $C(X)$, i.e.,
\begin{eqnarray*}
\widehat{R}_{(i/n)} \underset{p}{\longrightarrow}&& \Mean_{Z\in \mathbb{X}}[ C(Z) \mathbb{I}\{r_{(\alpha)} \leq C(Z) - \delta\}] -\delta =\Mean_{Z\in \mathbb{X}}\{C(Z)  | C(Z)\geq r_{(\alpha)} + \delta\} - \delta,
\end{eqnarray*}
where $r_{(\alpha)}+\delta$ is the upper $i/n$ quantile of the density of $C(X)$ when $n$ goes to infinity.  As long as $\widehat{R}_{(i/n)}$ is larger than zero, the selected subgroup satisfies the condition in \eqref{objective} based on the theoretical optimal SSR in \eqref{eqn:ossr2} from Theorem \ref{thm1}. Therefore, we need to select patients with positive $\widehat{R}_{(i/n)}$ and maximize the subgroup size to solve \eqref{objective}. To do this, we define the reward of the $i$-th individual based on the sign of $\widehat{R}_{(i/n)}$ as follows:

\textbf{Reward 1:}
\begin{eqnarray}\label{reward1}
\Gamma_i^{(1)} (D)=  \mathbb{I}\{D(X_i)=1\} \left[ \text{sign}\{\widehat{R}_{(K_i)}\}  \right],
\end{eqnarray} 
where $K_i$ is the rank of $\widehat{r}_i$ in the sequence $\{\widehat{r}_{(i)}\}$ or the sequence $\{\widehat{R}_{(i/n)}\}$, and `$ \text{sign}$' is the sign operator such that $\text{sign}\{x\}=1$ if $x> 0$, $\text{sign}\{x\}=0$ if $x= 0$, and $\text{sign}\{x\}=-1$ if $x< 0$. Given $\widehat{R}_{(K_i)}$ is positive, the reward $\Gamma_i^{(1)} $ is 1 if the patient is selected to be part of the subgroup, and is 0 otherwise. Likewise, supposing $\widehat{R}_{(K_i)}$ is negative, the reward $\Gamma_i^{(1)} $ is $-1$ if the patient is selected to be in the subgroup, i.e., $D(X_i)=1$, and is 0 otherwise. This is in accordance with the intuition that we should select patients with $\widehat{R}_{(K_i)}$ larger than zero.

To encourage the decision tree to include patients who have a lager treatment effect, we also propose the following reward choice based on the value of $\widehat{R}_{(K_i)}$ directly:

\textbf{Reward 2:}
\begin{eqnarray}\label{reward2}
\Gamma_i^{(2)} (D)=  \mathbb{I}\{D(X_i)=1\} \left\{ \widehat{R}_{(K_i)} \right\}.
\end{eqnarray} 


 
The optimal SSR is searched within the decision tree class $\Pi_{DT}$ to maximize the sum of the individual rewards defined in \eqref{reward1} or \eqref{reward2}. Specifically, the decision tree allocates each patient to the subgroup or not, and receives the corresponding rewards. We use the exhaustive search to estimate the optimal SSR that optimizes the total reward, using the policy tree algorithm proposed in Athey and Wager (2017) \citep{athey2021policy}. It is shown in the simulation studies (Section \ref{sec:simu}) that the performances are very similar under these two reward choices. 
We denote the estimated optimal SSR that maximizes the size of the subgroup and also maintains the desired average treatment effect as $\widehat{D}(\cdot)$. The proposed algorithm not only results in an interpretable SSR (see more discussion in Section \ref{sec:simu}), but also is flexible to handle multiple constraints and survival data, as discussed in detail in the next section.

 \section{Extensions}\label{sec:extens}
In this section, we discuss two main extensions of CAPITAL for solving \eqref{objective}. We first address multiple constraints on the average treatment effect in Section \ref{sec:ext1}, and then handle the time to event data with the restricted mean survival time as the clinically interesting mean outcome in Section \ref{sec:ext2}.
   \subsection{Extension to Multiple Constraints}\label{sec:ext1}
In addition to the main constraint described in \eqref{objective}, in reality there may exist secondary constraints of interest. For instance, besides a desired average treatment effect, the individual treatment effect for each patient should be greater than some minimum beneficial value. Under such multiple constraints, the optimal SSR is defined by
   \begin{eqnarray}\label{objective_multi}
	&&\max_{D\in \Pi}\quad   \prob \{D(X)=1\},\\\nonumber
	&&\text{s.t.} \quad \Mean\{Y^*(1)|D(X)=1\}-\Mean\{Y^*(0)|D(X)=1\} \ge \delta>0,\\\nonumber
	&&\text{s.t.} \quad \Mean\{Y^*(1)|D(X)=1,X=x\}-\Mean\{Y^*(0)|D(X)=1,X=x\} \ge \gamma, \forall x \in \mathbb{X},
\end{eqnarray}  
where $\gamma$ is a pre-specified minimum beneficial value. In the rest of this paper, we focus on the case with $\gamma=0$, that is, the individual treatment effect for each patient should be nonnegative so that the treatment is beneficial to the patients in the selected group.  \textcolor{black}{Following similar arguments in \eqref{objective}, we can derive the empirical version of the optimization objective in \eqref{objective_multi} as
\begin{eqnarray}\label{objective_ex1}
	&& \max_{D\in \Pi}\quad  \sum_{i=1}^n \mathbb{I}\{D(X_i)=1\}, \\\nonumber
&& \text{s.t.} \quad  \sum_{i=1}^n \widehat{C}(X_i) \mathbb{I}\{D(X_i)=1\} \ge n \delta>0, \quad \text{  and } \quad  \widehat{C}(X_i) \ge \gamma, \forall  i.
\end{eqnarray}}
The above objective function can be solved by modifying CAPITAL presented in Section \ref{sec:algo}. Specifically, we define the reward of the $i$-th individual based on \eqref{objective_ex1} and \eqref{reward2} as follows.

\textbf{Reward 3:}
\begin{eqnarray}\label{reward3}
\Gamma_i^{(3)} (D)=  \mathbb{I}\{D(X_i)=1\} \left[ \widehat{R}_{(K_i)} + \lambda  \mathbb{I}\{\widehat{C}(X_i)<0\} \widehat{C}(X_i) \right],
\end{eqnarray} 
where $\lambda$ is the nonnegative penalty parameter that represents the trade-off between the first  and the second constraint. When $\lambda=0$, the reward defined in \eqref{reward3} reduces to \eqref{reward2}.  Here, we only add the penalty on the reward when the estimated contrast function is negative, i.e., $ \mathbb{I}\{\widehat{C}(X_i)<0\}$. This prevents the method from selecting patients with a negative individual treatment effect.

  \subsection{Extension to Survival Data}\label{sec:ext2}
We next consider finding the optimal SSR for a survival endpoint. Let $T_i$ and $C_i$ denote the survival time of interest and the censoring time, respectively. Assume that $T_i$ and $C_i$ are independent given baseline covariates and the treatment. Then, the observed dataset consists of independent and identically distributed triplets, $\{(X_i, A_i,\tilde{T}_i, \Delta_i), i = 1, \cdots , n\}$, where $\tilde{T}_i= \min(T_i, C_i)$ and $\Delta_i= \mathbb{I}( T_i\leq C_i)$. The goal is to maximize the size of the subgroup with a pre-specified clinically desired effect $\delta$, i.e.,
\begin{eqnarray}\label{objective_survival}
	&& \max_{D\in \Pi}\quad \Mean \mathbb{I}\{D(X)=1\}, \\\nonumber &&\text{s.t.} \quad \Mean \{\min(T,\tau)|D(X)=1, A=1\}	-  \Mean\{\min(T,\tau)|D(X)=1, A=0 \}\ge \delta,
\end{eqnarray}
where $\tau$ is the maximum follow up time, which is pre-specified or can be estimated based on the observed data.  Denote $ {\mu}_0(X) = \int_0^\tau S(t|A=0)dt$ and ${\mu}_1(X)=  \int_0^\tau S(t|A=1)dt$ as the restricted mean survival time for groups with treatment 0 and 1, respectively, given baseline covariate $X$, where $S(t|A=0)$ and $S(t|A=1)$ are survival functions in the control and treatment groups, respectively. To estimate $ {\mu}_0(X)$ and ${\mu}_1(X)$, we first fit a random forest on the survival functions in the control and treatment groups, respectively, and \textcolor{black}{obtain estimates} as $\widehat{S}(t|A=0)$ and $\widehat{S}(t|A=1)$. Then, the estimated restricted mean survival time for groups with treatment 0 and 1, denoted as $\widehat{\mu}_0(X)$ and $\widehat{\mu}_1(X)$, are calculated by integrating the estimated survival functions to the minimum of the maximum times over the 2 arms. \textcolor{black}{Similarly, the empirical version of optimization in \eqref{objective_survival} is 
\begin{eqnarray}\label{objective_ex2}
	 \max_{D\in \Pi}\quad  \sum_{i=1}^n \mathbb{I}\{D(X_i)=1\}, \quad \quad \text{s.t.} \quad  \sum_{i=1}^n\{ \widehat{\mu}_1(X_i) - \widehat{\mu}_0(X_i) \}\mathbb{I}\{D(X_i)=1\} \ge n \delta>0.
\end{eqnarray}} 
Define $\widehat{r}_i = \widehat{\mu}_1(X_i)-\widehat{\mu}_0(X_i) -\delta$ to capture the distance from the estimated contrast function to the desired difference in restricted mean survival time $\delta$ for the $i$-th individual.  It is immediate that an individual with larger $\widehat{r}_i$ is more likely to be selected into the subgroup. We sort the estimates $\widehat{r}_i$ as $
\widehat{r}_{(1)}\geq\widehat{r}_{(2)}\geq\cdots\geq\widehat{r}_{(n)} $ and define the cumulative mean as $ \widehat{R}_{(i/n)} ={ i^{-1}}\sum_{j=1}^i \widehat{r}_{(j)}$. The reward for the constrained policy tree search can be defined following similar arguments as in \eqref{reward1} and \eqref{reward2}.

 \section{Simulation Studies} \label{sec:simu}
 

\subsection{Evaluation and Comparison with Average Treatment Effect}\label{sec:simu_ate}
\subsubsection{Data Generation}
 Suppose baseline covariates $X=[X^{(1)},\cdots,X^{(r)}]^\top$, the treatment information $A$, and the outcome $Y$ are generated from the following model:
\begin{equation}\label{gen_model} 
\begin{split}
&A \overset{iid}{\sim} \text{Bernoulli} \{0.5\}, \quad X^{(1)}, \cdots,X^{(r)} \overset{iid}{\sim} \text{Uniform}[-2,2],\\
&Y=U(X)+AC(X)+\epsilon,
\end{split}
\end{equation}
where $U(\cdot)$ is the baseline function of the outcome, $C(\cdot)$ is the contrast function, $\epsilon \overset{i i d}{\sim} N(0,1)$ is the random error. 
We set the dimension of covariates as $r=10$ and consider the following three scenarios respectively.

\noindent \textbf{Scenario 1}:   
$	 \begin{array}{ll}
		U(X)={X^{(1)}}+2X^{(2)},\quad C(X)=  X^{(1)} . 
	\end{array} $

\noindent \textbf{Scenario 2}: 
$
	\begin{array}{ll}
		U(X)={X^{(1)}}+2X^{(2)},\quad C(X)=X^{(1)}\times X^{(2)}.
	\end{array} $

\noindent \textbf{Scenario 3}: 
$
	 \begin{array}{ll}
		U(X)={X^{(1)}}+2X^{(2)},\quad
		C(X)=X^{(1)}-X^{(2)}. 
	\end{array} $

\begin{figure}[!t] 
 \centering
\begin{subfigure}{}
	\centering
	\includegraphics[width=0.45\textwidth]{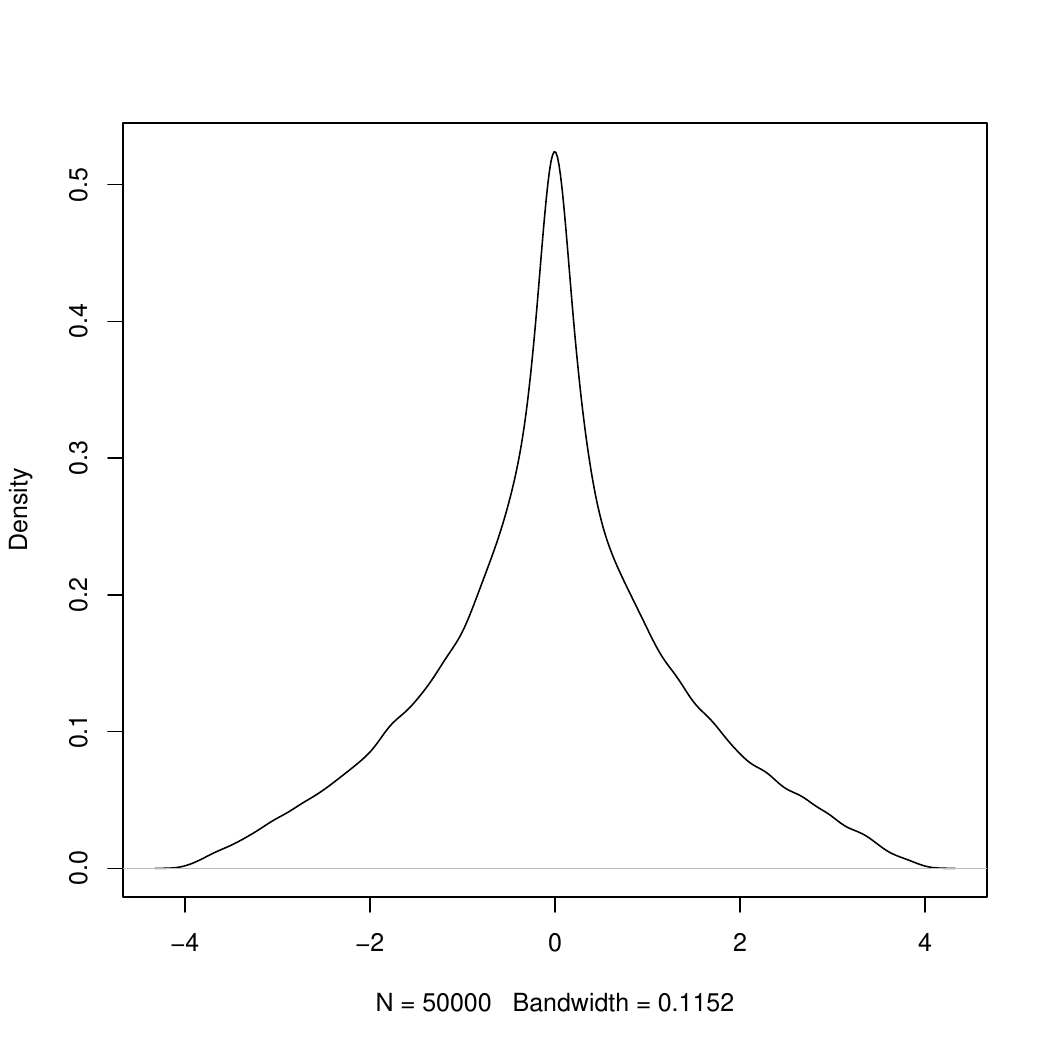}
\end{subfigure}
\begin{subfigure}{}
	\centering
	\includegraphics[width=0.45\textwidth]{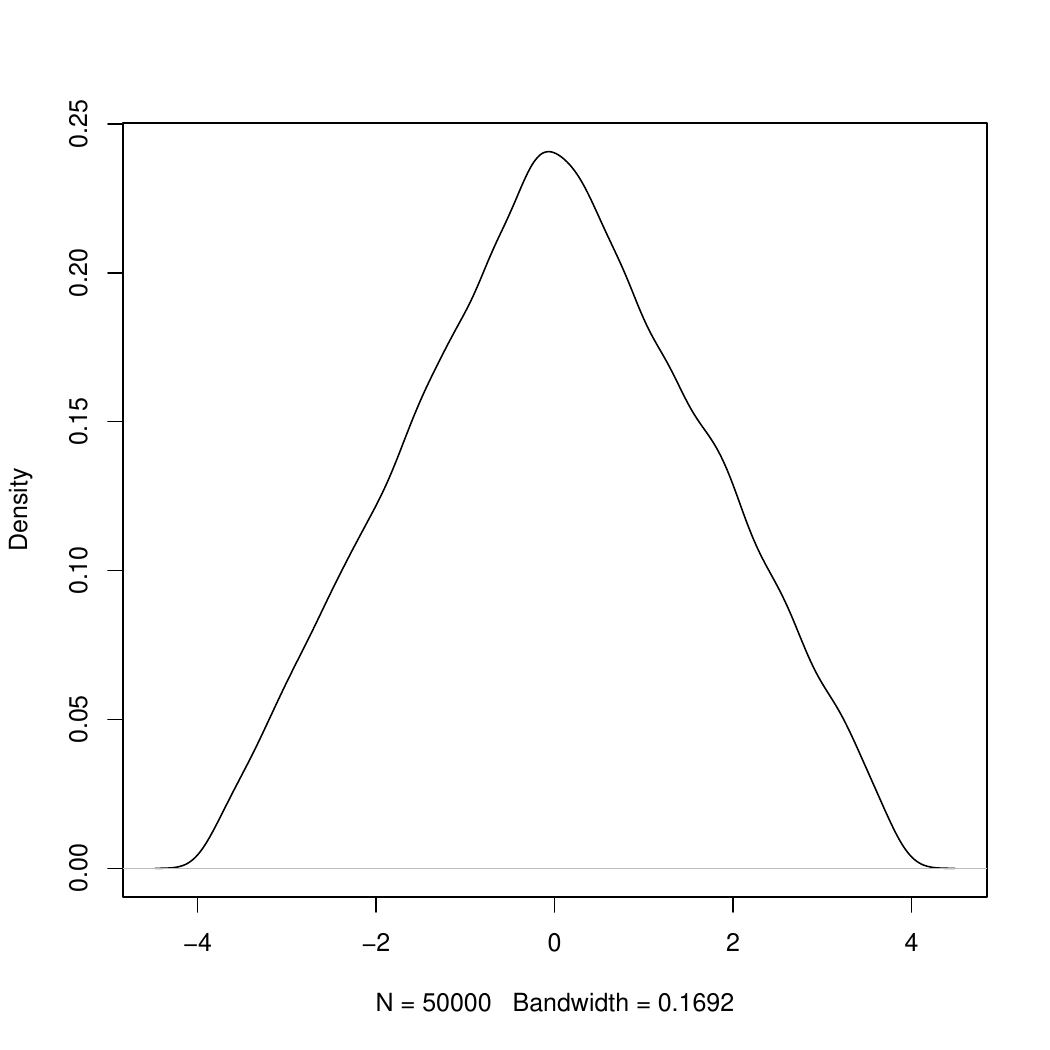}
\end{subfigure}
	\caption{Left panel: the density function of $C(X)$ for simulation Scenario 2. Right Panel: the density function of $C(X)$ for simulation Scenario 3.} \label{fig:dens}
\end{figure}

The true average treatment effect can be calculated as 0 under all scenarios. We illustrate the density of $C(X)$ for Scenarios 2 and 3 in Figure \ref{fig:dens}. Note the density of $C(X)$ for \textcolor{black}{Scenario 1} is just a uniform distribution on interval $[-2,2]$. Based on Figure \ref{fig:dens}, we consider the clinically meaningful treatment effect $\delta \in \{0.7,1.0,1.3\} $ for all scenarios, with the corresponding optimal subgroup sample proportions as listed in Table \ref{table:1}. Let the total sample size $n$ be chosen from the set $\{200, 500, 1000\}$. 

 \subsubsection{Results under CAPITAL}
 
\begin{figure}[!t]  
 \centering
\begin{subfigure}{}
	\centering
	\includegraphics[width=0.48\textwidth]{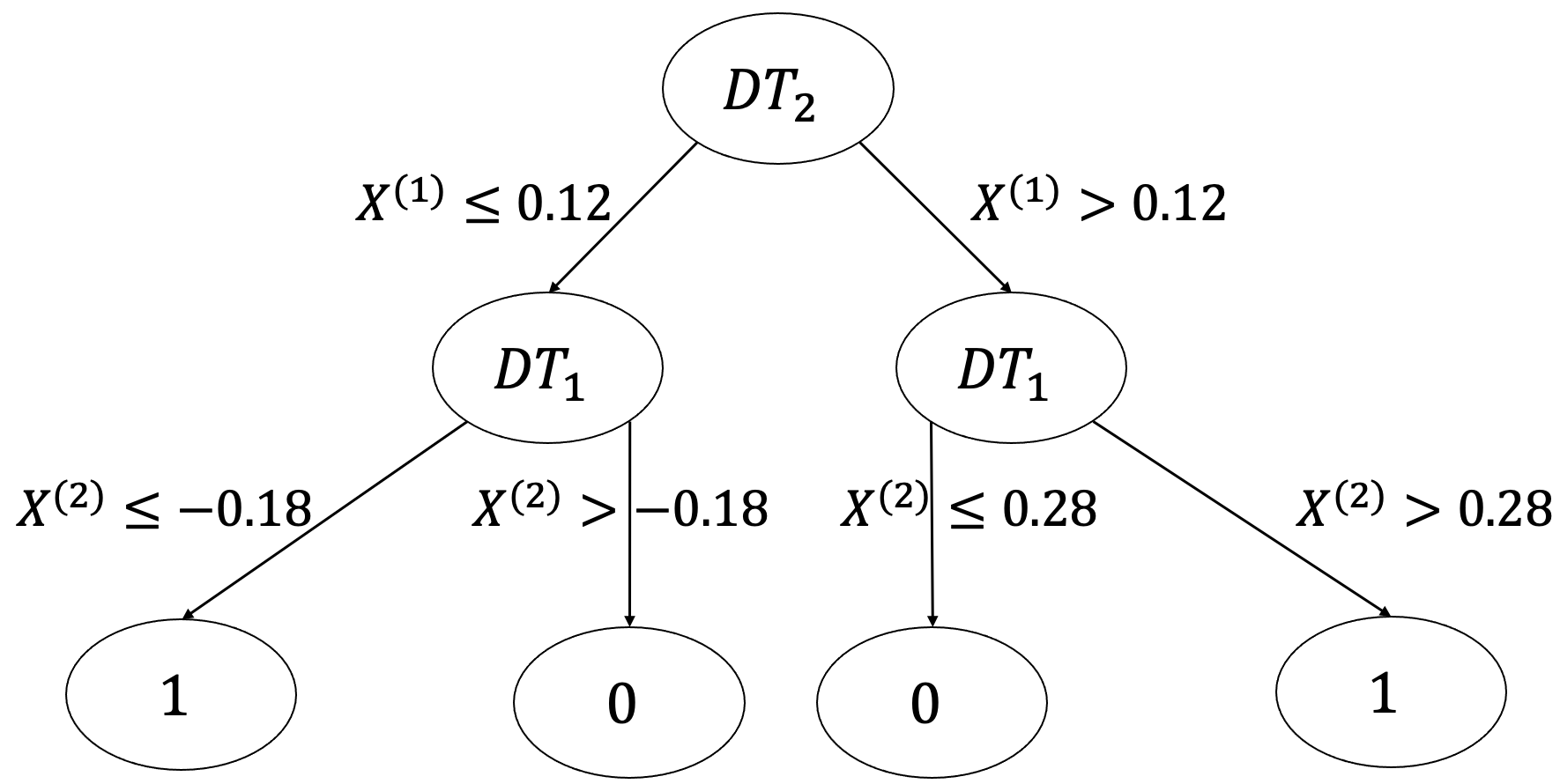}
\end{subfigure}
\begin{subfigure}{}
	\centering
	\includegraphics[width=0.48\textwidth]{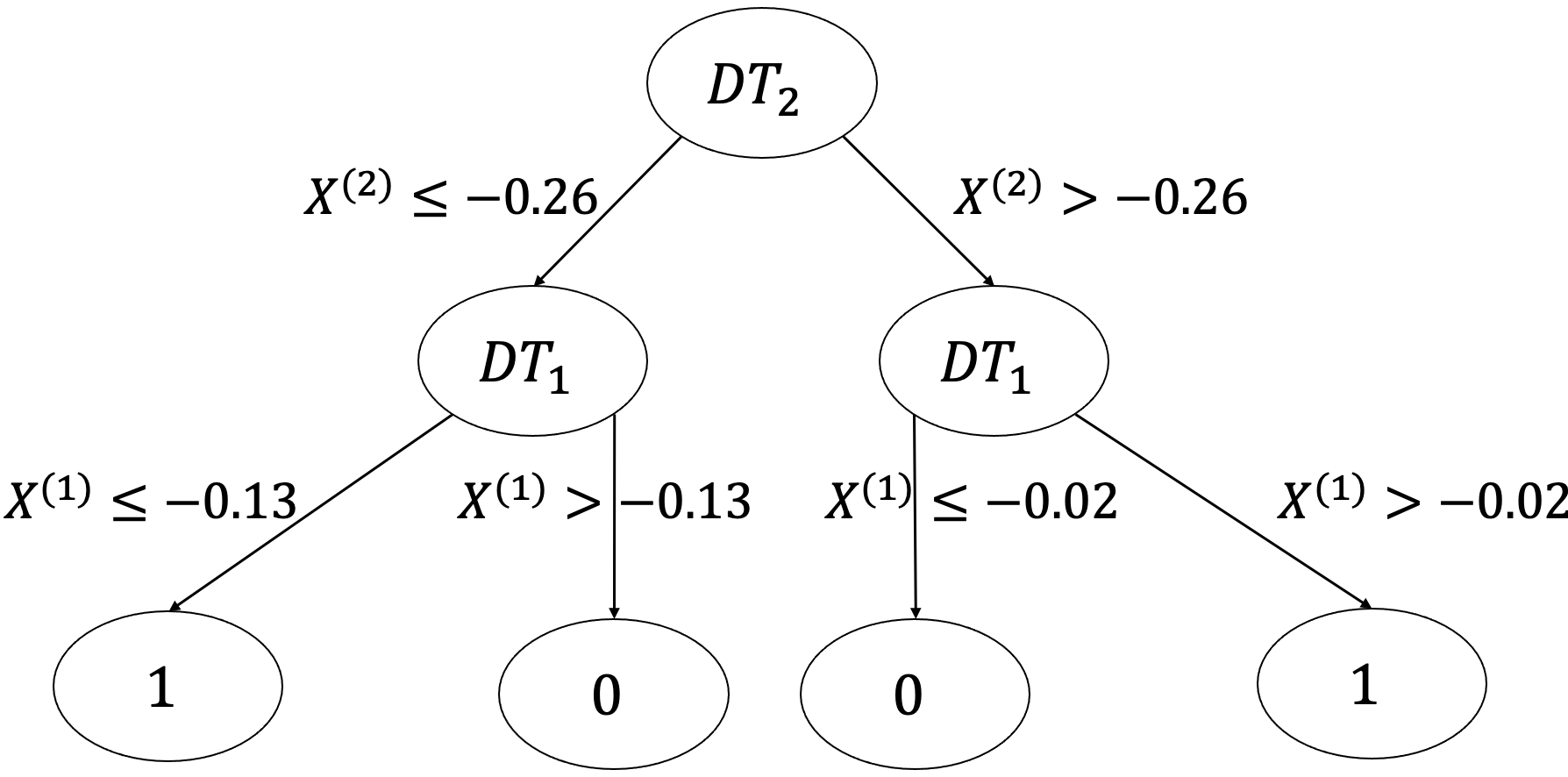}
\end{subfigure}
\begin{subfigure}{}
	\centering
	\includegraphics[width=0.48\textwidth]{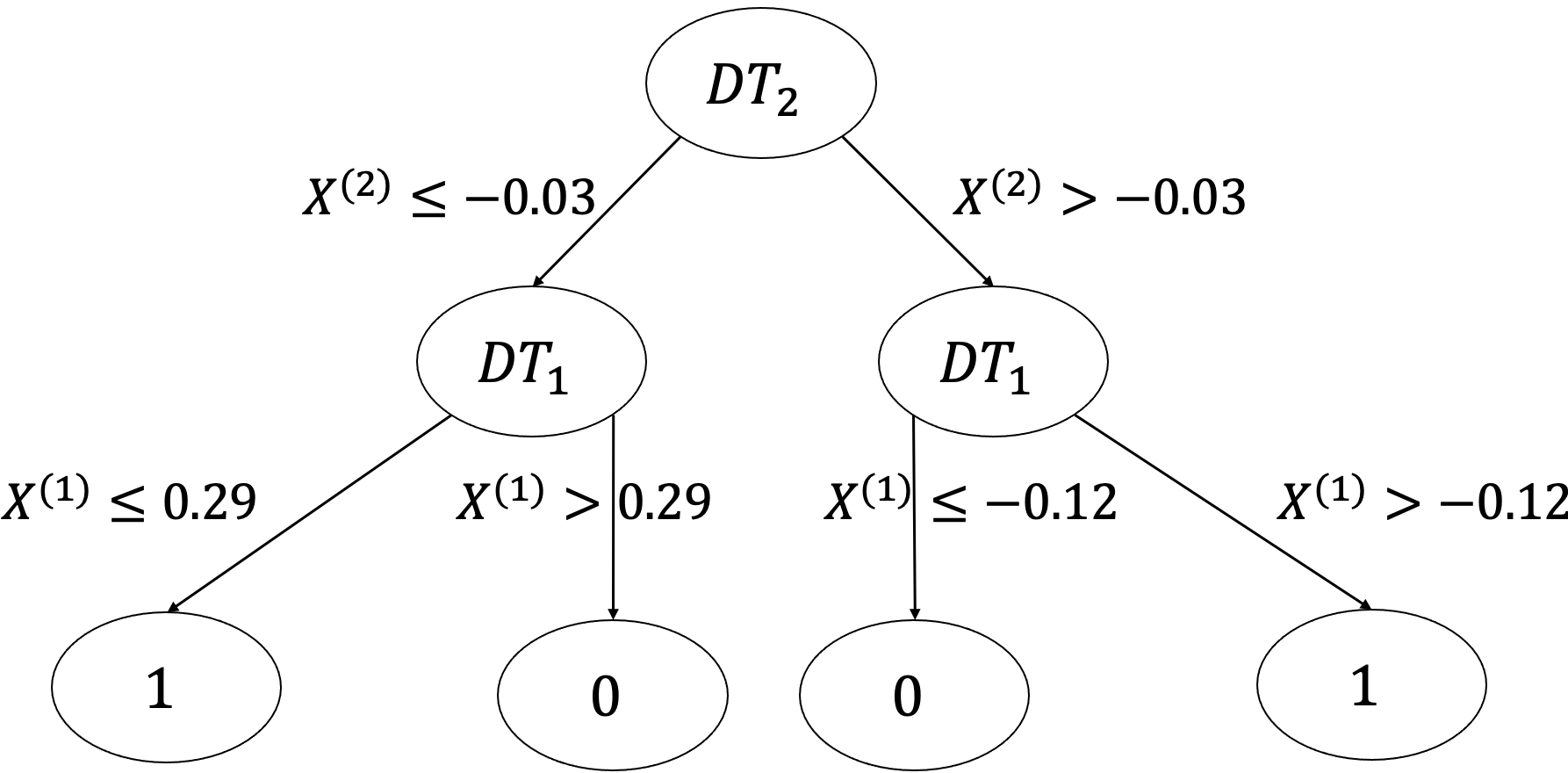}
\end{subfigure}
	\caption{The estimated optimal subgroup selection tree by CAPITAL under Scenario 2 with $\delta=1.0$ and $n=1000$. Upper left panel: for replicate No.1. Upper right Panel: for replicate No.2. Lower middle Panel: for replicate No.3.} 
	 \label{fig:show_dts}
\end{figure}

\begin{figure}[!t]
 \centering
\begin{subfigure}{}
	\centering
	\includegraphics[width=0.3\textwidth]{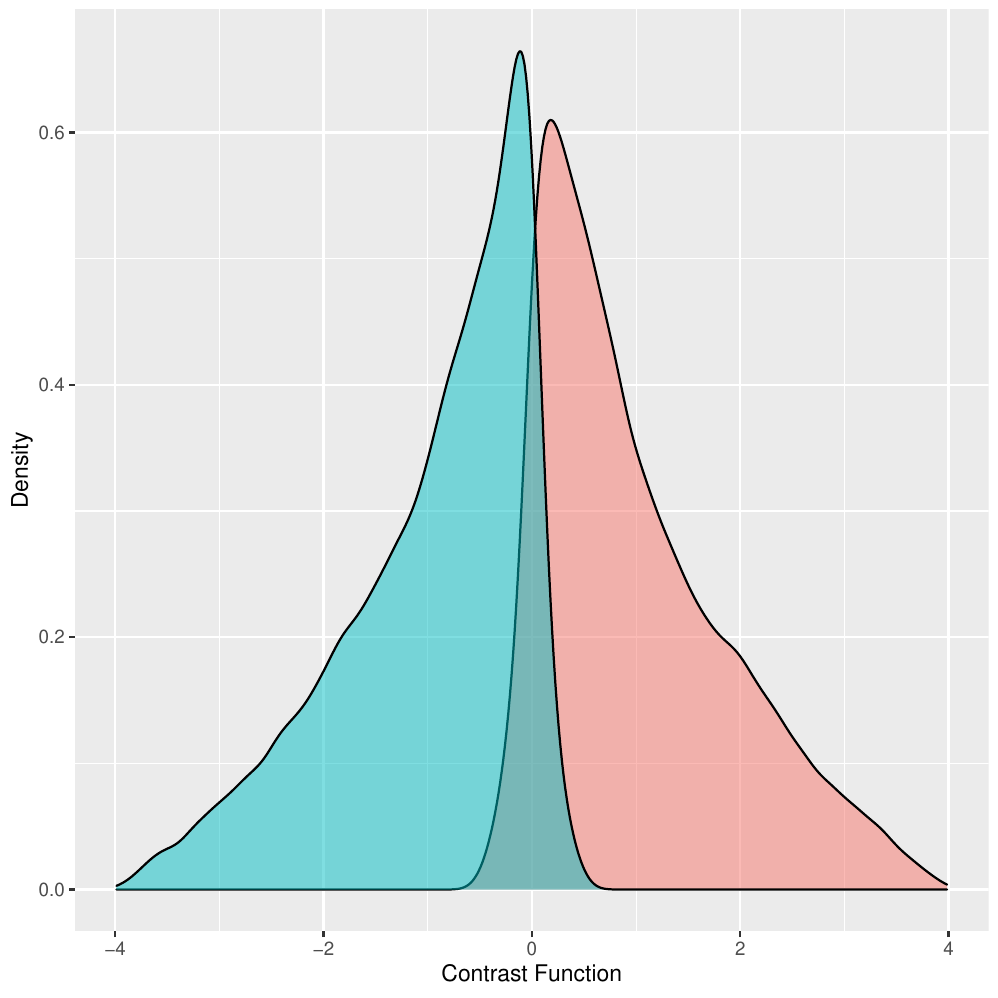}
\end{subfigure}
\begin{subfigure}{}
	\centering
	\includegraphics[width=0.3\textwidth]{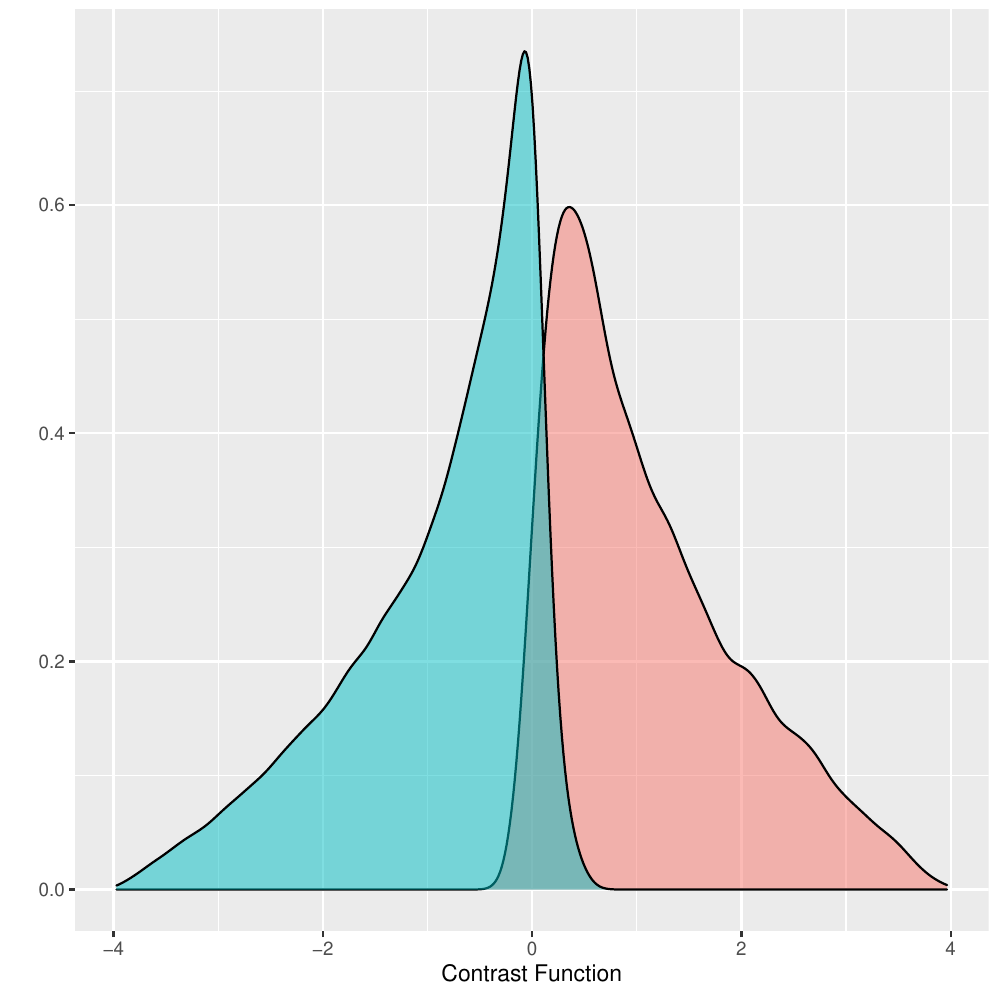}
\end{subfigure}
\begin{subfigure}{}
	\centering
	\includegraphics[width=0.33\textwidth]{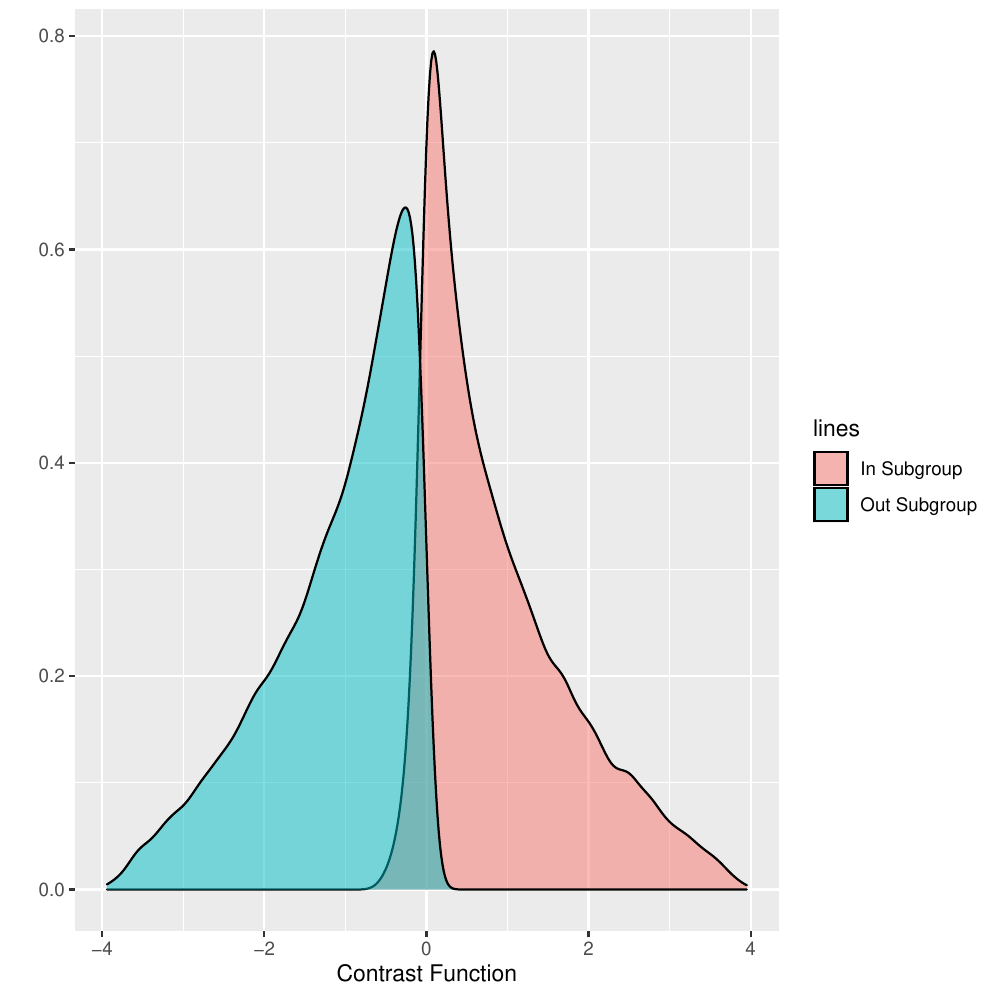}
\end{subfigure}
	\caption{The density function of $C(X)$ within or outside the subgroup under Scenario 2 with $\delta=1.0$ and $n=1000$. Left panel: for replicate No.1. Middle Panel: for replicate No.2. Right Panel: for replicate No.3.} 
	 \label{fig:show_cx}
\end{figure}

We apply CAPITAL to find the optimal SSR. The policy is searched within $\Pi_{DT}$ based on the \textsf{R} package `policytree' \citep{athey2021policy,zhou2018offline}. For better demonstration, we focus on $L=2$ decision trees. To illustrate the interpretability of the resulting SSR, we \textcolor{black}{highlight the results of three specific simulation replicates (denoted as replicate No.1, No.2, and No.3)} under Scenario 2 with $\delta=1.0$ using \textcolor{black}{the first reward \eqref{reward2}} for $n=1000$. The estimated SSR under these three selected replicates are shown in Figure \ref{fig:show_dts}, with the splitting variables and their splitting thresholds reported in Table \ref{table_toy}. We summarize the selected sample proportion $\prob\{\widehat{D}(X)\} $ under the estimated SSR, the average treatment effect $ {ATE}(\widehat{D}) $ of the estimated SSR, and the rate of making correct subgroup decisions by the estimated SSR, using Monte Carlo approximations. Finally, we visualize the density function of $C(X)$ within the subgroup selected by the estimated \textcolor{black}{SSR and compare to} that of unselected \textcolor{black}{patients for three} replicates in Figure \ref{fig:show_cx}.

Over 200 replicates \textcolor{black}{under Scenario 2 with $\delta=1.0$}, the rate of correctly identifying important features ${X^{(1)}}$ and $X^{(2)}$ under the estimated SSRs is 70.8\% with $n=200$, increasing to 95.8\% with $n=500$, and 100.0\% with $n=1000$. It can be seen from both Figure \ref{fig:show_dts} and Table \ref{table_toy} that the estimated SSRs under the proposed method \textcolor{black}{identify the important features that determine the outcome, ${X^{(1)}}$ and $X^{(2)}$, for all three replicates}. In Scenario 2, ${X^{(1)}}$ and $X^{(2)}$ have identical roles in the contrast function, so the resulting optimal tree can either use ${X^{(1)}}$ or $X^{(2)}$ as the first splitting variable.  Replicate No.3 over-selects the subgroup and therefore yields a lower average treatment effect, while replicate No.1 under-selects the subgroup and achieves a higher average treatment effect, as shown in Table \ref{table_toy}. This finding is in line with the trade-off between the size of the selected subgroup and its corresponding average treatment effect discussed in the introduction. Moreover, all these three replicates have a high rate ($> 90\%$) of making correct subgroup decisions under the estimated SSRs, supported by both Table \ref{table_toy} and Figure \ref{fig:show_cx}.

\begin{table}[!t]
\centering
\caption{Results of estimated optimal subgroup selection tree for three particular replicates under Scenario 2 with $\delta=1.0$ and $n=1000$ (where the optimal subgroup sample proportion is $50\%$) under CAPITAL.}\label{table_toy}
\scalebox{0.9}{
 \begin{tabular}{c|c|c|c}
\toprule
	Simulation &  Replicate No.1&Replicate No.2&Replicate No.3\\
\midrule
 $\prob\{\widehat{D}(X)\} $&44.5\%&49.2\%&55.0\%   \\
 	\midrule 
 	${ATE}(\widehat{D})$&1.11&1.00&0.90   \\
	 	\midrule 
 	Rate of Correct Decision &91.85\%&92.01\%&94.45\%  \\
	\midrule 
	$DT_2$ Split Variable (Split Value)& $X^{(1)}(0.12)$ & $X^{(2)}(-0.26)$& $X^{(2)}(-0.03)$\\
 	\midrule
	$DT_1$(Left) Split Variable (Split Value)& $X^{(2)}(-0.18)$ & $X^{(1)}(-0.13)$& $X^{(1)}(0.29)$\\
 	\midrule
	$DT_1$(Right) Split Variable (Split Value)& $X^{(2)}(0.28)$ & $X^{(1)}(-0.02)$& $X^{(1)}(-0.12)$\\
		\bottomrule
\end{tabular}}
\end{table}

 \subsubsection{Comparison Studies}
 \textcolor{black}{We compare the proposed method with two popular methods with variations: two variants of the VT method \citep{foster2011subgroup} and two variants of the policy tree search method \citep{fu2016estimating} using the adjusted value estimator. While the VT method can
theoretically be used for both binary and continuous outcomes, the current \textsf{R} package `aVirtualTwins' only deals with binary outcomes in a two-armed clinical trial. To address the continuous outcomes in Scenarios 1-3, following the VT method \citep{foster2011subgroup} we fit the estimated individual treatment effect $ \widehat{C}(X) $ on features via a regression tree. We consider two subgroup selection rules based on the VT method.}

\textbf{VT-A:} Denote the average treatment effect within a terminal node $T$ as $\widehat{Z}(T)={|T|^{-1}}\sum_{j\in T} \widehat{C}(X_j) $ \textcolor{black}{ where $|\cdot|$ is the size of the terminal node}. The final subgroup is formed as the union of the terminal nodes where the predicted values $\widehat{Z}(T)$ are greater than $\delta$. 

\textbf{VT-C:} Denote $u_i = \mathbb{I}(\widehat{C}(X_i)>\delta)$. Then each terminal node $T$ is classified into the subgroup based on a majority vote within the node by $\widehat{U}(T)\equiv \mathbb{I}\{{|T|}^{-1}\sum_{i\in T} u_i >0.5\}$. The final subgroup is defined as the union of the terminal nodes with $\widehat{U}(T)= 1$. 

 \textcolor{black}{The second method considered for comparison finds subgroups by maximizing the $\delta$-quality adjusted value estimator  \citep{fu2016estimating,bai2017optimal} via a policy tree search. Specifically, we treat $Y-\delta$ or $C(X)-\delta$ as the individual value and select the $i$-th individual into the treatment benefitting subgroup if its value is positive. The optimal decision tree can be found by maximizing $\delta$-quality adjusted value estimators, i.e., $\argmax_{D\in \Pi}\quad  \sum_{i=1}^n D(X_i)(Y_i-\delta) $ or  $\argmax_{D\in \Pi}\quad  \sum_{i=1}^n D(X_i)\{\widehat{C}(X_i)-\delta\}$ 
 for the $Y-\delta$ or $C(X)-\delta$ individual rewards, respectively \citep{fu2016estimating,bai2017optimal}.}
  
 \textcolor{black}{ We apply the proposed method, the VT-A and VT-C methods, and the policy tree search methods based on different $\delta$-quality adjusted value estimators, with 200 replications on Scenarios 1-3. Under the estimated SSR, we summarize the following estimates and corresponding standard deviations, aggregated using Monte Carlo approximations over 200 replications in Table \ref{table:1}: the selected sample proportion as $\prob\{\widehat{D}(X)\} $, the average treatment effect as $ {ATE}(\widehat{D}) $, the rate of making correct subgroup decisions (RCD, the number of correct subgroup decisions divided by the total sample size), and the rate of positive individual treatment effect within the selected subgroup (RPI, the number of positive individual treatment effects divided by the size of the selected subgroup).
 Since the performance of our method under reward \eqref{reward1} and reward \eqref{reward2} are similar, and the VT-A and VT-C methods have nearly identical results, we do not report the proposed method with reward \eqref{reward2} and the VT-A method in the empirical results Table \ref{table:1}. These results can be found in Table \ref{table:2} in the appendix. 
 The results of policy tree search by different $\delta$-quality adjusted value estimators are included in Table \ref{table:tree_com2} in the appendix using Scenario 1 as an illustration. }

\begin{sidewaystable}[!thp] 
\centering
\caption{Empirical results of subgroup analysis under the estimated optimal SSR by CAPITAL with reward in \eqref{reward1} and the VT-C method.
}\label{table:1}
\scalebox{0.7}{
 \begin{tabular}{cccccc|ccc|ccc}
 	\toprule
 	
 	Method&&$r=10$&\multicolumn{3}{c|}{Scenario 1} &\multicolumn{3}{c|}{Scenario 2} &\multicolumn{3}{c}{Scenario 3}\\
 	\cmidrule{3-12}
 	&&& $n=200$ & $n=500$&$n=1000$& $n=200$ & $n=500$&$n=1000$& $n=200$ & $n=500$&$n=1000$\\
\toprule
 	CAPITAL&$\delta=0.7$ &Proportion&&$65\%$&&&$67\%$&&&$75\%$&\\
 	\cmidrule{2-12} 
 &&$\prob\{\widehat{D}(X)\} $ &0.62(0.16) & 0.63(0.08) & 0.65(0.05) & 0.42(0.23) & 0.51(0.11) & 0.56(0.05) & 0.72(0.15) & 0.74(0.08) & 0.77(0.05) \\
 	\cmidrule{3-12}
 	&&${ATE}(\widehat{D})$&  0.66(0.28) & 0.72(0.17) & 0.69(0.10) & 0.72(0.47) & 0.96(0.20) & 0.86(0.11) & 0.66(0.34) & 0.67(0.18) & 0.61(0.11) \\
	\cmidrule{3-12}
 	&&RCD&0.83(0.10) & 0.91(0.05) & 0.93(0.03) & 0.62(0.15) & 0.81(0.08) & 0.87(0.03) & 0.83(0.08) & 0.89(0.03) & 0.90(0.01) \\
	 \cmidrule{3-12}
 	&&RPI&0.78(0.13) & 0.80(0.09) & 0.78(0.06) & 0.74(0.15) & 0.88(0.08) & 0.86(0.06) & 0.67(0.10) & 0.67(0.06) & 0.65(0.04) \\
	\cmidrule{2-12} 
 	&$\delta=1.0$ &Proportion&&$50\%$&&&$50\%$&&&$63\%$&\\
 	\cmidrule{2-12} 
	&&$\prob\{\widehat{D}(X)\} $ &0.46(0.16) & 0.48(0.09) & 0.50(0.06) & 0.21(0.17) & 0.32(0.12) & 0.40(0.06) & 0.56(0.16) & 0.59(0.09) & 0.62(0.06) \\
 	\cmidrule{3-12}
 	&&${ATE}(\widehat{D})$&  0.90(0.27) & 1.00(0.15) & 0.99(0.11) & 0.83(0.63) & 1.31(0.27) & 1.17(0.11) & 1.02(0.37) & 1.00(0.20) & 0.94(0.15) \\
	\cmidrule{3-12}
 	&&RCD&0.84(0.11) & 0.91(0.05) & 0.94(0.03) & 0.62(0.12) & 0.79(0.11) & 0.88(0.05) & 0.79(0.07) & 0.85(0.03) & 0.87(0.01) \\
	 \cmidrule{3-12}
 	&&RPI&0.88(0.11) & 0.94(0.06) & 0.94(0.05) & 0.75(0.19) & 0.95(0.05) & 0.97(0.03) & 0.78(0.11) & 0.78(0.06) & 0.77(0.05) \\
	\cmidrule{2-12} 
 	&$\delta=1.3$ &Proportion&&$35\%$&&&$37\%$&&&$51\%$&\\
 	\cmidrule{2-12} 
	&&$\prob\{\widehat{D}(X)\} $ &0.30(0.16) & 0.31(0.11) & 0.34(0.08) & 0.09(0.09) & 0.14(0.10) & 0.25(0.09) & 0.41(0.15) & 0.44(0.09) & 0.48(0.06) \\
 	\cmidrule{3-12}
 	&&${ATE}(\widehat{D})$& 1.05(0.33) & 1.28(0.18) & 1.29(0.14) & 0.66(0.73) & 1.58(0.59) & 1.48(0.24) & 1.36(0.40) & 1.38(0.24) & 1.29(0.15) \\
	\cmidrule{3-12}
 	&&RCD&0.81(0.10) & 0.88(0.07) & 0.92(0.04) & 0.67(0.07) & 0.74(0.08) & 0.82(0.06) & 0.78(0.08) & 0.83(0.03) & 0.86(0.02) \\
	 \cmidrule{3-12}
 	&&RPI&0.93(0.12) & 0.99(0.02) & 1.00(0.01) & 0.69(0.21) & 0.91(0.13) & 0.95(0.04) & 0.86(0.10) & 0.89(0.06) & 0.88(0.04) \\
 \toprule
 	VT-C&$\delta=0.7$ &Proportion&&$65\%$&&&$67\%$&&&$75\%$&\\
\cmidrule{2-12} 
 &&$\prob\{\widehat{D}(X)\} $ &0.31(0.12) & 0.34(0.09) & 0.35(0.08) & 0.15(0.10) & 0.19(0.09) & 0.22(0.08) & 0.29(0.10) & 0.30(0.06) & 0.30(0.06)  \\
 	\cmidrule{3-12} 
 	&&${ATE}(\widehat{D})$&  1.11(0.20) & 1.27(0.17) & 1.30(0.15) & 0.85(0.61) & 1.46(0.38) & 1.53(0.32) & 1.76(0.36) & 1.82(0.23) & 1.81(0.21)  \\
	\cmidrule{3-12}
 	&&RCD&0.66(0.12) & 0.69(0.09) & 0.70(0.08) & 0.43(0.08) & 0.51(0.09) & 0.55(0.09) & 0.54(0.10) & 0.55(0.06) & 0.55(0.06)  \\
	 \cmidrule{3-12}
 	&&RPI&0.97(0.06) & 0.99(0.03) & 1.00(0.01) & 0.77(0.17) & 0.95(0.09) & 0.97(0.08) & 0.95(0.07) & 0.98(0.03) & 0.98(0.03)  \\
\cmidrule{2-12} 
 	&$\delta=1.0$ &Proportion&&$50\%$&&&$50\%$&&&$63\%$&\\
\cmidrule{2-12} 
	&&$\prob\{\widehat{D}(X)\} $ &0.21(0.13) & 0.24(0.10) & 0.26(0.07) & 0.07(0.06) & 0.09(0.07) & 0.14(0.07) & 0.23(0.09) & 0.24(0.06) & 0.25(0.05)  \\
 	\cmidrule{3-12}
 	&&${ATE}(\widehat{D})$&  1.19(0.21) & 1.37(0.18) & 1.45(0.13) & 1.01(0.74) & 1.67(0.49) & 1.78(0.38) & 1.94(0.34) & 2.02(0.23) & 2.00(0.18)  \\
	\cmidrule{3-12}
 	&&RCD& 0.70(0.12) & 0.74(0.10) & 0.76(0.07) & 0.54(0.06) & 0.59(0.07) & 0.64(0.07) & 0.60(0.08) & 0.62(0.06) & 0.62(0.05)  \\
	 \cmidrule{3-12}
 	&&RPI&0.98(0.05) & 1.00(0.02) & 1.00(0.00) & 0.81(0.20) & 0.96(0.09) & 0.98(0.08) & 0.97(0.05) & 0.99(0.02) & 0.99(0.01)  \\
\cmidrule{2-12} 
 	&$\delta=1.3$ &Proportion&&$35\%$&&&$37\%$&&&$51\%$&\\
\cmidrule{2-12} 
	&&$\prob\{\widehat{D}(X)\} $ &0.12(0.11) & 0.11(0.11) & 0.16(0.11) & 0.03(0.04) & 0.03(0.04) & 0.07(0.05) & 0.17(0.09) & 0.18(0.06) & 0.20(0.05)  \\
 	\cmidrule{3-12}
 	&&${ATE}(\widehat{D})$&1.25(0.23) & 1.43(0.18) & 1.50(0.12) & 1.11(0.81) & 1.81(0.61) & 1.98(0.42) & 2.12(0.37) & 2.24(0.23) & 2.19(0.20)  \\
	\cmidrule{3-12}
 	&&RCD& 0.74(0.09) & 0.76(0.11) & 0.81(0.11) & 0.65(0.03) & 0.66(0.04) & 0.69(0.05) & 0.65(0.09) & 0.67(0.06) & 0.69(0.05)  \\
	 \cmidrule{3-12}
 	&&RPI&0.99(0.04) & 1.00(0.01) & 1.00(0.00) & 0.83(0.21) & 0.95(0.13) & 0.98(0.07) & 0.99(0.03) & 1.00(0.01) & 1.00(0.00)  \\
	 	\bottomrule
\end{tabular}}
\end{sidewaystable}

\textcolor{black}{Based on Tables \ref{table:1}, \ref{table:2}, and \ref{table:tree_com2}, it is clear that the proposed method has a better performance than the VT methods and the $\delta$-quality adjusted policy tree search methods in all cases. In Scenario 1 with} $n=1000$, our method achieves a selected sample proportion of 65\% for $\delta=0.7$ (the optimal is 65\%), 50\% for $\delta=1.0$ (the optimal is 50\%), and 34\% for $\delta=1.3$ (the optimal is 35\%), with corresponding average treatment effects close to the true values. \textcolor{black}{In Scenario 2, the selected sample proportion of the proposed method is moderately underestimated }due to the fact that the density function of $C(X)$ is concentrated around \textcolor{black}{0, as illustrated in the left panel of Figure \ref{fig:dens}. 
The proposed method performs well under small sample sizes, but with slightly lower  selected sample proportion, and gets better performance in selected subgroups} as the sample size increases. \textcolor{black}{In contrast, in most cases, the VT methods identify subgroups that are barely half of the desired optimal subgroup size. According to Table \ref{table:tree_com2}, simply using $Y-\delta$ or $C(X)-\delta$ as the individual reward in policy tree with $\delta$-quality adjusted estimator method \citep{fu2016estimating,bai2017optimal} cannot achieve the desired optimal subgroup proportion and underestimate/overestimate the average treatment effect. The results with $C(X)-\delta$ as the individual value are very close to the results under the VT-C method, where we use $\widehat{C}(X_i)-\delta$ as the individual reward with the majority vote for subgroup tree search. Though the policy tree can find a relatively larger subgroup when using $Y-\delta$ as the individual value, the corresponding average treatment effects are much smaller than both our proposed method and the VT methods in all cases.} 

   \subsection{Evaluation of Multiple Constraints}
 In this section, we further investigate the performance of the proposed method under multiple constraints. Specifically, we aim to solve the objective in \eqref{objective_ex1} with the penalized reward defined in \eqref{reward3}. \textcolor{black}{We define four cases based on penalty terms $\lambda\in \{0,0.5,1,2\}$, where $\lambda=0$ corresponds to \eqref{reward2}, i.e., a single constraint. We use the same setting as described in Section \ref{sec:simu_ate}  with $\delta = 0.7 $ under Scenarios 1 to 3 and apply CAPITAL to find the optimal SSR within $\Pi_{DT}$. The empirical results are reported in Table \ref{table:4} under the different penalty terms $\lambda$ over 200 replications. It can be observed from Table \ref{table:4} that for all cases, as the penalty term $\lambda$ increases, the rate of positive individual treatment effect within the selected subgroup increases while the rate of making correct subgroup decisions slightly decreases.} This reflects the trade-off between two constraints in \textcolor{black}{our theoretical objective in \eqref{objective_multi}}.

\begin{sidewaystable}[!thp] 
\centering 
\caption{Empirical results of optimal subgroup selection tree by CAPITAL with the penalized reward in \eqref{reward3}.}\label{table:4}
\scalebox{0.9}{
 \begin{tabular}{ccccc|ccc|ccc}
\toprule
 	&$r=10$&\multicolumn{3}{c|}{Scenario 1} &\multicolumn{3}{c|}{Scenario 2} &\multicolumn{3}{c}{Scenario 3}\\
 	\cmidrule{3-11}
 	&& $n=200$ & $n=500$&$n=1000$& $n=200$ & $n=500$&$n=1000$& $n=200$ & $n=500$&$n=1000$\\
\midrule
 	$\delta=0.7$ &Proportion&&$65\%$&&&$67\%$&&&$75\%$&\\
\midrule
$\lambda=0$ &$\prob\{\widehat{D}(X)\} $ &0.63(0.16) & 0.63(0.08) & 0.65(0.05) & 0.44(0.24) & 0.51(0.11) & 0.57(0.06) & 0.72(0.15) & 0.75(0.07) & 0.77(0.04) \\
 	\cmidrule{2-11}
 	&${ATE}(\widehat{D})$&  0.67(0.30) & 0.72(0.17) & 0.70(0.11) & 0.71(0.48) & 0.95(0.20) & 0.85(0.11) & 0.67(0.35) & 0.66(0.17) & 0.60(0.10) \\
	\cmidrule{2-11}
 	&RCD&0.84(0.10) & 0.91(0.05) & 0.93(0.03) & 0.62(0.15) & 0.81(0.08) & 0.87(0.03) & 0.83(0.08) & 0.89(0.03) & 0.91(0.01) \\
	 \cmidrule{2-11}
 	&RPI&0.78(0.13) & 0.80(0.09) & 0.78(0.06) & 0.74(0.16) & 0.88(0.09) & 0.85(0.07) & 0.67(0.10) & 0.67(0.06) & 0.65(0.04) \\
\midrule
$\lambda=0.5$ &$\prob\{\widehat{D}(X)\} $ &0.55(0.12) & 0.56(0.06) & 0.57(0.04) & 0.39(0.21) & 0.48(0.10) & 0.53(0.05) & 0.63(0.13) & 0.65(0.07) & 0.66(0.05) \\
 	\cmidrule{2-11}
 	&${ATE}(\widehat{D})$&  0.83(0.23) & 0.86(0.11) & 0.86(0.08) & 0.77(0.48) & 1.01(0.17) & 0.93(0.10) & 0.89(0.30) & 0.88(0.16) & 0.86(0.11) \\
	\cmidrule{2-11}
 	&RCD&0.84(0.09) & 0.90(0.05) & 0.91(0.03) & 0.61(0.15) & 0.79(0.08) & 0.85(0.04) & 0.81(0.09) & 0.86(0.04) & 0.87(0.03) \\
	 \cmidrule{2-11}
 	&RPI&0.86(0.11) & 0.88(0.07) & 0.88(0.05) & 0.76(0.15) & 0.91(0.07) & 0.90(0.05) & 0.74(0.09) & 0.74(0.06) & 0.74(0.04) \\
\midrule
$\lambda=1$ &$\prob\{\widehat{D}(X)\} $ &0.52(0.11) & 0.54(0.05) & 0.54(0.04) & 0.37(0.20) & 0.46(0.09) & 0.51(0.05) & 0.57(0.13) & 0.60(0.07) & 0.61(0.05) \\
 	\cmidrule{2-11}
 	&${ATE}(\widehat{D})$&  0.88(0.20) & 0.91(0.11) & 0.91(0.07) & 0.79(0.48) & 1.05(0.16) & 0.97(0.10) & 1.00(0.29) & 0.99(0.16) & 0.98(0.12) \\
	\cmidrule{2-11}
 	&RCD&0.83(0.09) & 0.88(0.05) & 0.89(0.04) & 0.60(0.15) & 0.78(0.08) & 0.83(0.05) & 0.78(0.10) & 0.83(0.05) & 0.84(0.04) \\
	 \cmidrule{2-11}
 	&RPI&0.88(0.09) & 0.90(0.06) & 0.91(0.05) & 0.77(0.15) & 0.92(0.06) & 0.92(0.05) & 0.78(0.09) & 0.78(0.05) & 0.78(0.04) \\
\midrule
$\lambda=2$ &$\prob\{\widehat{D}(X)\} $ &0.49(0.11) & 0.52(0.05) & 0.52(0.04) & 0.33(0.19) & 0.43(0.09) & 0.48(0.05) & 0.52(0.12) & 0.55(0.07) & 0.55(0.05) \\
 	\cmidrule{2-11}
 	&${ATE}(\widehat{D})$&  0.93(0.19) & 0.95(0.11) & 0.96(0.07) & 0.83(0.51) & 1.10(0.15) & 1.03(0.10) & 1.12(0.30) & 1.11(0.16) & 1.11(0.12) \\
	\cmidrule{2-11}
 	&RCD&0.81(0.10) & 0.86(0.05) & 0.87(0.04) & 0.58(0.15) & 0.76(0.08) & 0.81(0.05) & 0.74(0.10) & 0.78(0.06) & 0.79(0.04) \\
	 \cmidrule{2-11}
 	&RPI&0.91(0.09) & 0.92(0.06) & 0.94(0.05) & 0.78(0.16) & 0.94(0.05) & 0.94(0.04) & 0.81(0.09) & 0.82(0.05) & 0.83(0.04) \\
 	\bottomrule
	\end{tabular}}
\end{sidewaystable}

  \subsection{Evaluation of Survival Data}\label{sec:simu_surv}
The data is generated by a similar model \textcolor{black}{as} \eqref{gen_model}:
  \begin{equation*} 
\begin{split}
 A \overset{iid}{\sim} \text{Bernoulli}(0.5), \quad X^{(1)},\cdots,X^{(r)} \overset{iid}{\sim} \text{Uniform}[-1,1], \quad Y=U(X)+AC(X)+\epsilon. 
\end{split}
\end{equation*}
We set the dimension of covariates as $r=10$, and define the survival time as $T = \exp(Y)$.  Consider the following scenario:

\noindent \textbf{Scenario 4}: 
$
	  \begin{array}{ll}
		U(X)=0.1{X^{(1)}}+0.2X^{(2)}, \quad C(X)=  X^{(1)} .
	\end{array} $

\begin{figure}[!t] 
	\centering
	\includegraphics[width=0.6\textwidth]{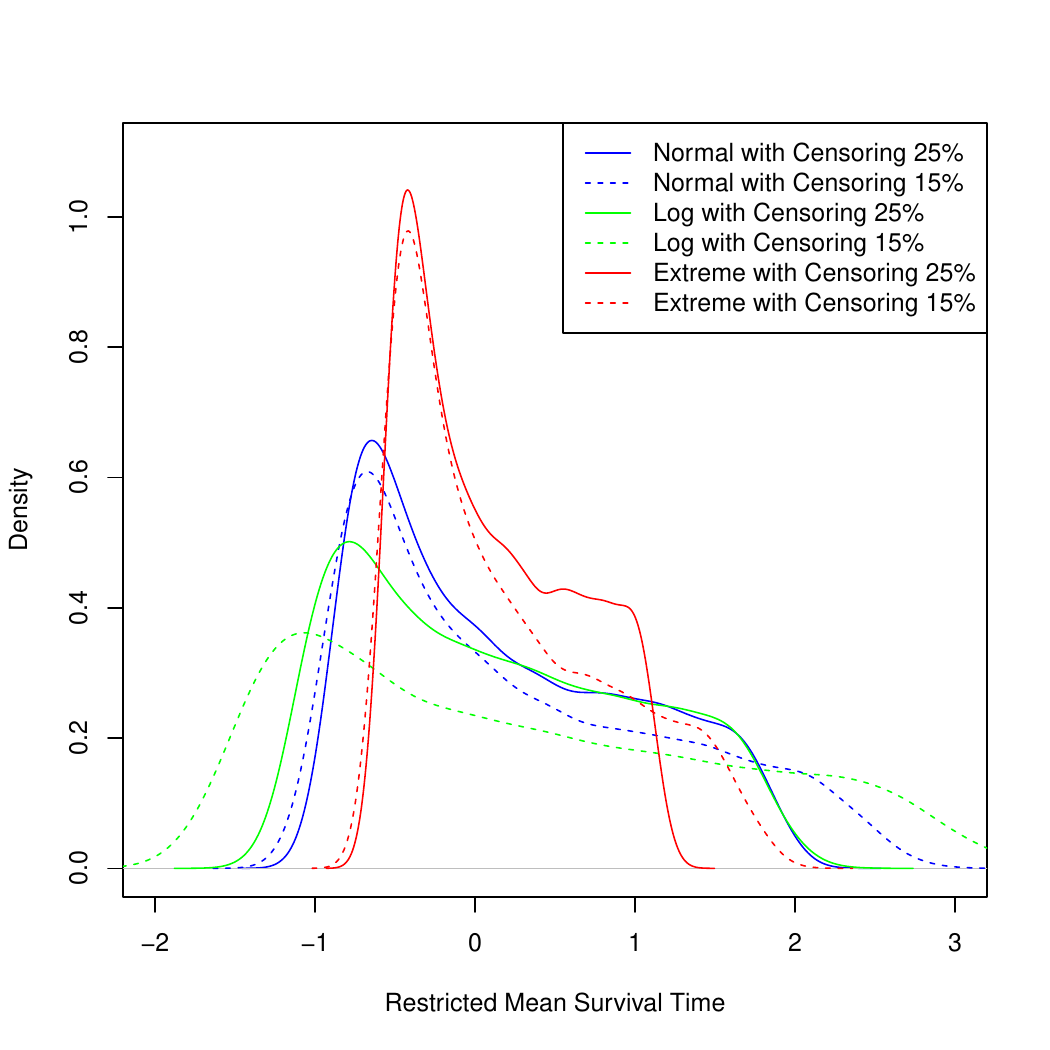}  
	\caption{The density function of restricted mean survival time ($ \Mean \{\min(T,\tau)| A=1\}	-  \Mean\{\min(T,\tau)| A=0 \}$) for Scenario 4 under different noises and censoring levels.} \label{fig:dens_survival}
\end{figure}

Here, for the random noise component we consider three cases: \\
(i) Case 1 (normal): $\epsilon\overset{i i d}{\sim} N(0,1)$; \\
(ii) Case 2 (logistic): $\epsilon \overset{i i d}{\sim} \text{logistic}(0,1)$; \\
(iii) Case 3 (extreme): $\epsilon \overset{i i d}{\sim} \log[-\log\{\text{Uniform}(0,1)\}]$.\\
The censoring times are generated from a uniform distribution on $[0, c_0]$, where $c_0$ is chosen to yield the desired censoring level 15\% and 25\%, respectively, each applied for the three choices of noise distributions for a total of 6 settings considered. We illustrate $ \Mean \{\min(T,\tau)| A=1\}	- \Mean\{\min(T,\tau)| A=0 \}$ in Figure \ref{fig:dens_survival}. The clinically meaningful difference in restricted mean survival time is summarized in Table \ref{table:3}. Each setting was selected to yield a selected sample proportion of 50\%. We report the empirical results in Table \ref{table:3} with the second choice of reward \eqref{reward2}, including the selected sample proportion $\prob\{\widehat{D}(X)\} $ under the estimated SSR, the average treatment effect $ {ATE}(\widehat{D}) $ of the estimated SSR, and the rate of making correct subgroup decisions by the estimated SSR, \textcolor{black}{over 200 replications using Monte Carlo approximations with standard deviations presented in parentheses. }

 \begin{table}[!t] 
\centering
\caption{Empirical results of optimal subgroup selection tree by CAPITAL for the survival data under Scenario 4 (where the optimal subgroup sample proportion is $50\%$).}\label{table:3}
\scalebox{0.9}{
 \begin{tabular}{cccc|cc}
\toprule
 	& &\multicolumn{2}{c|}{ Censoring Level $15\%$}  &\multicolumn{2}{c}{Censoring Level $25\%$}\\
 	\cmidrule{3-6}
 	&&  $n=500$&$n=1000$&     $n=500$&$n=1000$\\
\midrule
Case 1 (normal) &True $\delta$&\multicolumn{2}{c}{1.07} &\multicolumn{2}{c}{0.86}   \\
\cmidrule{2-6}  
&$\prob\{\widehat{D}(X)\} $&0.45(0.17)&0.47(0.12)&0.46(0.16) &0.48(0.11)   \\
 	\cmidrule{2-6} 
 	&${ATE}(\widehat{D})$&1.07(0.31)&1.11(0.24)&0.87(0.22) &0.87(0.16)   \\
	\cmidrule{2-6} 
	&RCD&0.84(0.11)&0.88(0.07)&0.84(0.09) &0.90(0.06)   \\
\midrule
 	Case 2 (logistic) &True $\delta$&\multicolumn{2}{c}{1.34} &\multicolumn{2}{c}{0.87}   \\
\cmidrule{2-6}  
&$\prob\{\widehat{D}(X)\} $&0.57(0.26)&0.56(0.18)&0.52(0.24) &0.52(0.18)   \\
 	\cmidrule{2-6} 
 	&${ATE}(\widehat{D})$&0.94(0.49)&1.06(0.36)&0.63(0.31) &0.75(0.24)   \\
	\cmidrule{2-6} 
	&RCD&0.72(0.13)&0.80(0.10)&0.74(0.13) &0.82(0.09)   \\
\midrule
 	Case 3 (extreme) &True $\delta$&\multicolumn{2}{c}{0.73} &\multicolumn{2}{c}{0.54}   \\
\cmidrule{2-6}  
&$\prob\{\widehat{D}(X)\} $&0.44(0.18)&0.46(0.12)&0.41(0.18) &0.44(0.12)   \\
 	\cmidrule{2-6} 
 	&${ATE}(\widehat{D})$&0.76(0.21)&0.78(0.15)&0.57(0.15) &0.58(0.11)   \\
	\cmidrule{2-6} 
	&RCD&0.84(0.11)&0.89(0.08)&0.83(0.12) &0.88(0.08)   \\
\bottomrule
\end{tabular}}
\end{table}

Table \ref{table:3} shows that the proposed method performs reasonably well under all three considered noise distributions. Both the selected sample proportion and average treatment effect under the estimated SSR get closer to the truth, and the rate of making correct subgroup decisions increases as the sample size increases. The selected sample proportion is slightly underestimated for Cases 1 and 3 where $C(X)$ has a more concentrated density function, while it is marginally overestimated for Case 2 \textcolor{black}{where the density function of $C(X)$ is more spread out.} All these findings are in accordance with our conclusions in Section \ref {sec:simu_ate}.

\section{Real Data Analysis}\label{sec:real}

In this section, we illustrate our proposed method by application to the AIDS Clinical Trials Group Protocol 175 (ACTG 175) data as described in  Hammer et al. (1996) \citep{hammer1996trial} and a Phase III clinical trial in patients with hematological malignancies from Lipkovich et al. (2017) \citep{lipkovich2017tutorial}.

\subsection{Case 1: ACTG 175 Data}\label{sec:real_acgt}
  
There were 1046 HIV-infected subjects enrolled in ACTG 175, randomized to two competing antiretroviral regimens \citep{hammer1996trial}: zidovudine (ZDV)+zalcitabine (zal) (denoted as treatment 0), and ZDV+didanosine (ddI) (denoted as treatment 1). Patients were randomized in equal proportions, with 524 patients randomized to treatment 0 and 522 patients to treatment 1, with constant propensity score  $\pi (x)\equiv 0.499$.  We consider $r=12$ baseline covariates: 1) four continuous variables: age (years), weight (kg), CD4 count (cells/mm3) at baseline, and CD8 count (cells/mm3) at baseline; and 2) eight categorical variables: hemophilia (0=no, 1=yes), homosexual activity (0=no, 1=yes), history of intravenous drug use (0=no, 1=yes), Karnofsky score (4 levels on the scale of 0-100, as 70, 80, 90, and 100), race (0=white, 1=non-white), gender (0=female), antiretroviral history (0=naive, 1=experienced), and symptomatic status (0=asymptomatic). The outcome of interest ($Y$) is the CD4 count (cells/mm3) at 20 $\pm$ 5 weeks. A higher CD4 count usually indicates a stronger immune system.  We normalize $Y$ by its mean and standard deviation. Our goal is to find the optimal subgroup selection rule that optimizes the size of the selected subgroup and achieves the desired average treatment effect.
 
       \begin{figure}[!t] 
	\centering
	\includegraphics[width=0.5\textwidth]{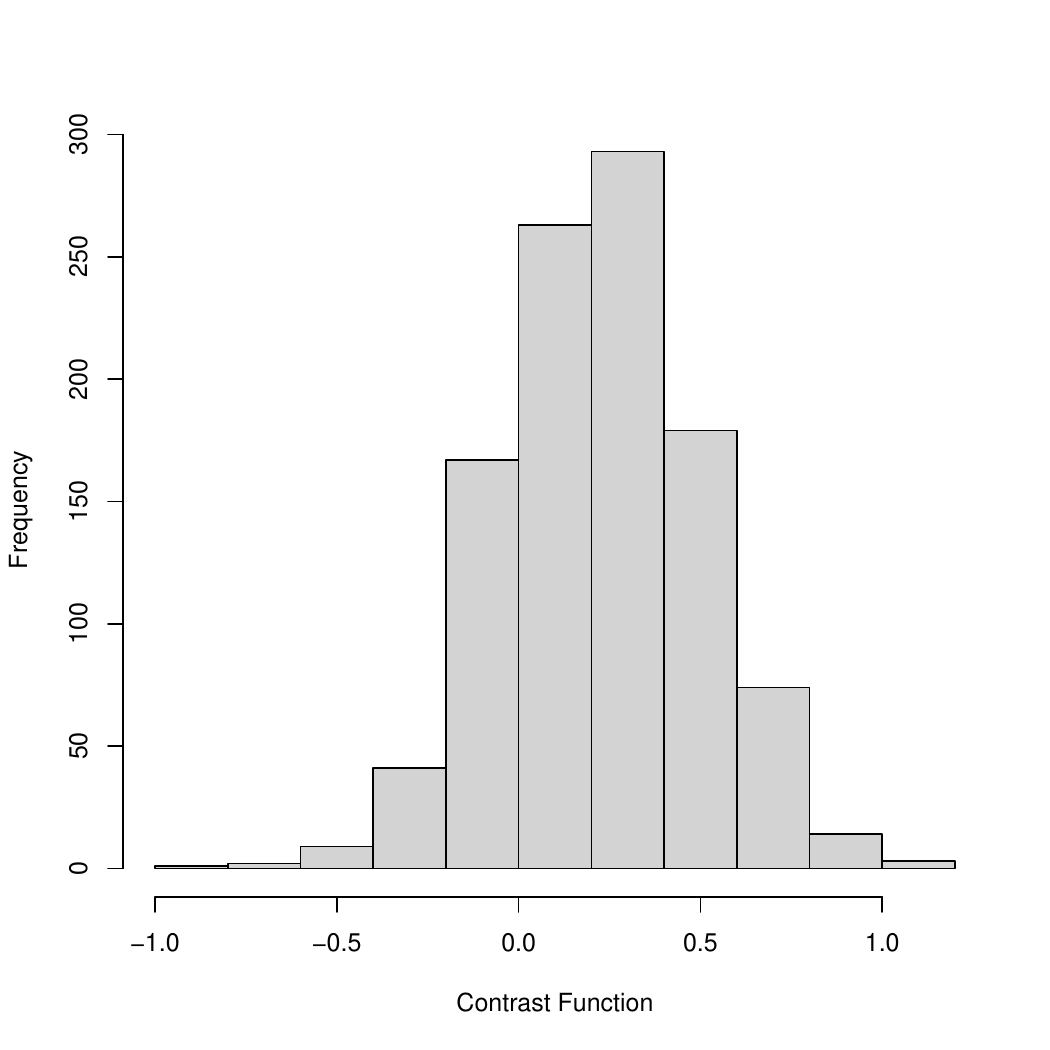} 
	\caption{The density function of the estimated contrast function $\widehat{C}(X)$  for the ACTG 175 data.} \label{fig:actg_den}
\end{figure}  
 
\begin{figure}[!t] 
 \centering
\begin{subfigure}{}
	\centering
	\includegraphics[width=0.45\textwidth]{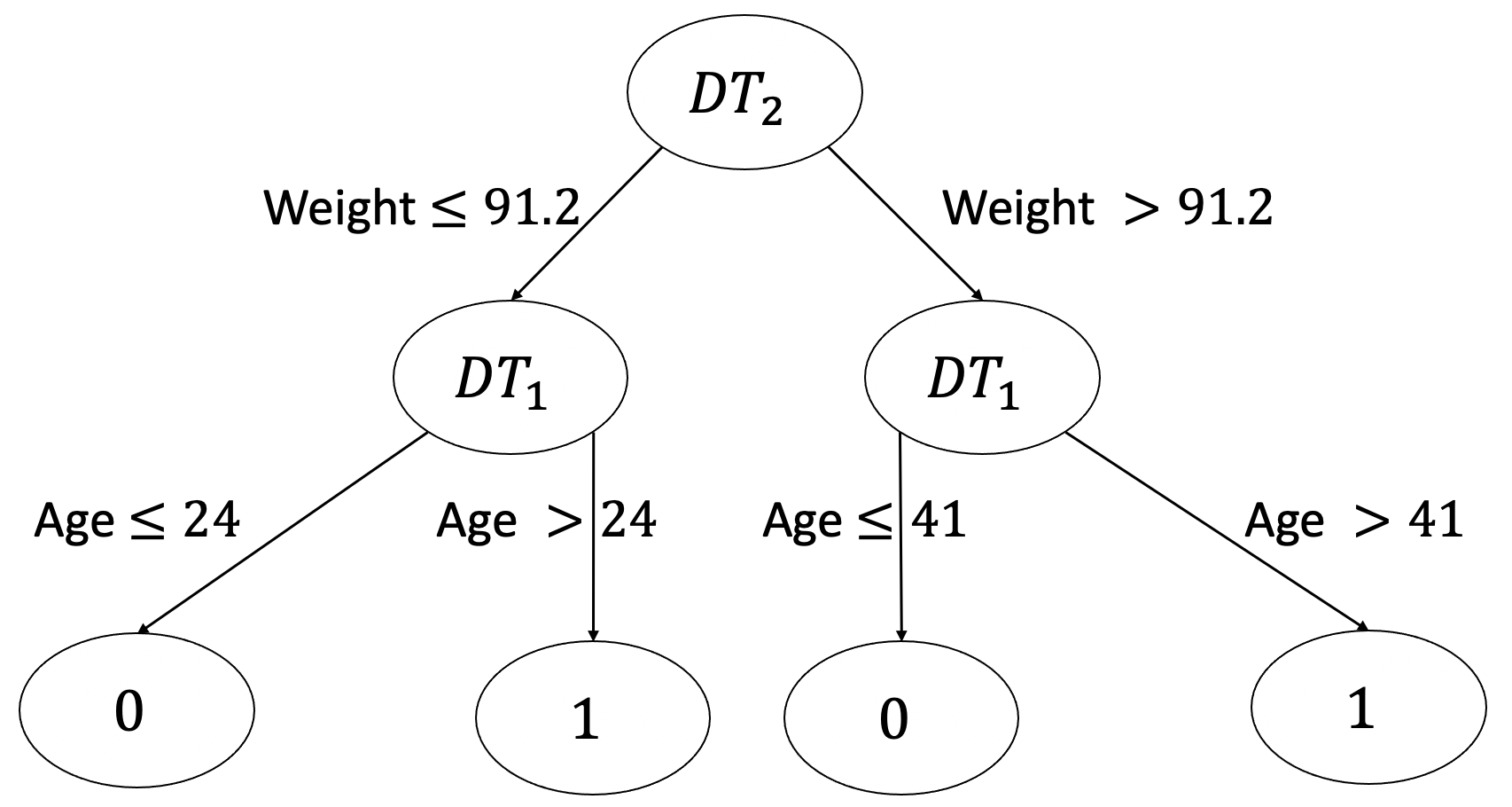}
\end{subfigure}
\begin{subfigure}{}
	\centering
	\includegraphics[width=0.47\textwidth]{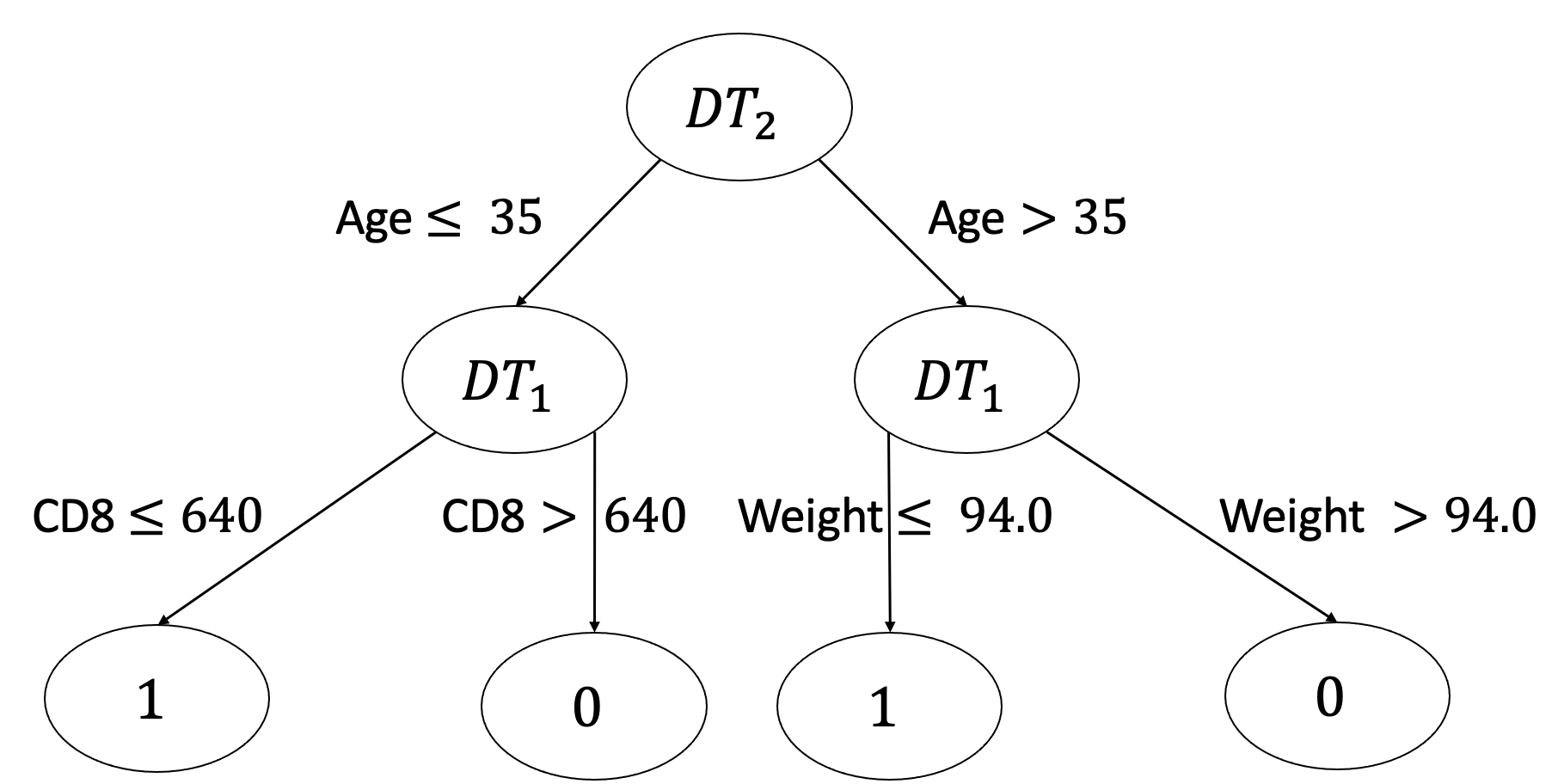}
\end{subfigure} 
	\caption{The estimated optimal subgroup selection tree using CAPITAL under the ACTG 175 data. Left panel: for $\delta=0.35$. Right Panel: for $\delta=0.40$.} 
	 \label{fig:show_dts_actg}
\end{figure}

 \textcolor{black}{In real analysis, the thresholds of average treatment effects, $\delta$, should be specified to exceed the whole group estimated average treatment effects by some extent while staying below the maximum individual treatment effects, to allow for identification of a reasonably sized benefiting subgroup. In practice, one can specify any clinical meaningful average treatment effect as the threshold and apply our method. In this example we use the estimated contrast density, along with the estimated mean contrast difference of 0.228, to select thresholds $\delta = 0.35$ and $\delta = 0.40$, corresponding to optimal subgroup sample proportions of approximately 40\% and 70\%, respectively.
 The density of the estimated contrast function $\widehat{C}(X)$ for the ACTG 175 data is provided in Figure \ref{fig:actg_den}.} 
We apply the proposed CAPITAL method and the virtual twin method (VT-C) \citep{foster2011subgroup}  \textcolor{black}{(because the results under the VT-C and VT-A methods as well as the $\delta$-quality adjusted policy tree search \citep{fu2016estimating,bai2017optimal} have nearly identical performances, as shown in the simulation studies)}, using the same procedure as described in Section \ref{sec:simu_ate}. The estimated SSRs under the proposed method are shown in Figure \ref{fig:show_dts_actg}. To evaluate the proposed method and VT-C method in the ACTG 175 data, we randomly split the whole data, with 70\% of the data as a training sample to find the SSR and 30\% as a testing sample to evaluate its performance. Here, we consider CAPITAL without penalty, with small penalty, and with large penalty on negativity of average treatment effect, respectively. The penalty term $\lambda$ is chosen from $ \{0, 4,20,100\}$, where $\lambda \in\{4,20,100\} $ encourages a positive average treatment effect in the selected group.  \textcolor{black}{Here, the magnitude of the (non-zero) penalty terms is selected based on the size of the average treatment effects.} In Table \ref{tab:actg}, we summarize the selected sample proportion $\prob\{\widehat{D}(X)\} $, the average treatment effect under the estimated SSR $ {ATE}(\widehat{D}) $, the average treatment effect outside the subgroup $ {ATE}(\widehat{D}^c) $, the difference of the average treatment effect within the subgroup and outside the subgroup $ {ATE}(\widehat{D}) - {ATE}(\widehat{D}^c) $, 
and the rate of positive individual treatment effect within the selected subgroup (RPI), aggregated over 200 replications with standard deviations presented in parentheses, under different $\delta $ for two methods.   

\begin{table}[!t] 
\centering
\caption{Evaluation results of the subgroup optimization using CAPITAL and the subgroup identification (using Virtual Twins \citep{foster2011subgroup}) under the ACTG 175 data.}\label{tab:actg}
\scalebox{0.9}{
 \begin{tabular}{cc|c|c}
\toprule
 	& Threshold& $\delta=0.35$ & $\delta=0.40$ \\ 
	\midrule
CAPITAL &$\prob\{\widehat{D}(X)\} $&92.8\% (0.023)&82.8\% (0.029)  \\
 	\cmidrule{2-4} 
 	without penalty &${ATE}(\widehat{D})$&0.250 (0.015)&0.270 (0.016)  \\
		 	\cmidrule{2-4} 
 	&${ATE}(\widehat{D}^c)$&-0.107 (0.069)&0.004 (0.038)\\
		 	\cmidrule{2-4} 
 	&${ATE}(\widehat{D})-{ATE}(\widehat{D}^c)$&0.357 (0.068)&0.266 (0.038)\\
	\cmidrule{2-4} 
	&RPI&83.0\% (0.021)&85.1\% (0.022)   \\
\midrule
CAPITAL &Penalty $\lambda=$ &4 &20 \\
 	\cmidrule{2-4} 
 	with small penalty  &$\prob\{\widehat{D}(X)\} $&52.7\% (0.052)&34.2\% (0.034)  \\
	\cmidrule{2-4} 
	&${ATE}(\widehat{D})$&0.327 (0.022)&0.385 (0.021)   \\
		 	\cmidrule{2-4} 
 	&${ATE}(\widehat{D}^c)$& 0.113 (0.021)&0.142 (0.017)\\
		 	\cmidrule{2-4} 
 	&${ATE}(\widehat{D})-{ATE}(\widehat{D}^c)$&0.214 (0.027)&0.243 (0.026)\\
	\cmidrule{2-4} 
	&RPI&91.5\% (0.029)&96.2\% (0.017)  \\
\midrule
CAPITAL &Penalty $\lambda=$ &20 & 100 \\
 	\cmidrule{2-4} 
 	with large penalty  &$\prob\{\widehat{D}(X)\} $&35.6\% (0.035)&19.5\% (0.051)\\
 	\cmidrule{2-4} 
 	 &${ATE}(\widehat{D})$&0.381 (0.021)&0.414 (0.032)  \\
		 	\cmidrule{2-4} 
 	&${ATE}(\widehat{D}^c)$&0.139 (0.017)&0.180 (0.017)\\
		 	\cmidrule{2-4} 
 	&${ATE}(\widehat{D})-{ATE}(\widehat{D}^c)$&0.242 (0.025)&0.234 (0.033)\\
	\cmidrule{2-4} 
	&RPI&95.9\% (0.017)&96.9\% (0.025)   \\ 	
\midrule
Virtual Twins &$\prob\{\widehat{D}(X)\} $&22.1\% (0.063)&10.5\% (0.029)  \\
 	\cmidrule{2-4} 
      &${ATE}(\widehat{D})$&0.462 (0.043)&0.556 (0.050)  \\
		 	\cmidrule{2-4} 
 	&${ATE}(\widehat{D}^c)$&0.159 (0.021)&0.187 (0.014)\\
		 	\cmidrule{2-4} 
 	&${ATE}(\widehat{D})-{ATE}(\widehat{D}^c)$&0.302 (0.037)&0.368 (0.047)\\
		\cmidrule{2-4} 
	&RPI&97.8\% (0.019)&99.6\% (0.010)   \\
\bottomrule
\end{tabular}}
\end{table}

As illustrated in Figure \ref{fig:show_dts_actg}, the estimated SSRs based on the proposed method under both $\delta = 0.35$ and $\delta = 0.40$ rely on the weight and age of patients. For instance, for a desired average treatment effect of 0.35, younger patients ($\leq 24$ years old) who weigh less than 91.2 kg or those $\leq 41$ years old weighting $>$ 91.2 kg may not benefit from treatment 1 (ZDV+ddI) and thus are not selected in the subgroup, while those older should be included into the subgroup of patients with enhanced effects from treating with ZDV+ddI. From Table \ref{tab:actg}, it is clear that the selected sample proportion under our method is much larger than that under the  \textcolor{black}{VT method in all cases. 
Specifically, our method yields a selected sample proportion at $92.8\%$ for $\delta=0.35$, and at $82.8\%$ for $\delta=0.40$, with a single constraint}. Under a penalty on negativity of average treatment effect, the size of the \textcolor{black}{identified subgroup using the proposed method is reduced to $73.4\%$ with small penalty $\lambda=4$ and to $35.6\%$ with large penalty  $\lambda=20$ under $\delta=0.35$, and further decreases to $34.2\%$} with small penalty $\lambda=20$ and to $19.5\%$ with large penalty  $\lambda=100$ under $\delta=0.40$. With a large penalty, our proposed method can achieve the desired average treatment effect at 0.381 (versus $\delta=0.35$) and at 0.414 (versus $\delta=0.40$). In contrast, the VT method identifies less than a quarter of the patients (22.1\%) in the case of $\delta=0.35$, and nearly a tenth of patients for $\delta=0.40$, with overestimated average treatment effects of 0.462 and 0.556, respectively. 
These imply that the proposed method could largely increase the number of benefitting patients to be selected in the subgroup while also maintaining the desired clinically meaningful threshold.


\subsection{Case 2: Phase III Trial for Hematological Malignancies}\label{sec:real_lip}
 
Next, we consider a Phase III randomized clinical trial in 599 patients with hematological malignancies \citep{lipkovich2017tutorial}. 
We exclude 7 subjects with missing records and use the remaining 592 complete records, for a final analysis dataset consisting of 301 patients receiving the experimental therapy plus best supporting care (as treatment 1) and 291 patients only receiving the best supporting care  (as treatment 0). We use the same $r=14$ baseline covariates selected by Lipkovich et al. (2017) \citep{lipkovich2017tutorial}: 1) twelve categorical variables: gender (1=Male, 2=Female), race (1= Asian, 2=Black, 3=White),  Cytogenetic markers 1 through 9 (0=Absent, 1=Present), and outcome for patient's prior therapy (1=Failure, 2=Progression, 3=Relapse); and 2) two ordinal variables: Cytogenetic category (1=Very good, 2=Good, 3 =Intermediate, 4=Poor, 5=Very poor),  and prognostic score for myelodysplastic syndromes risk assessment (IPSS) (1=Low, 2=Intermediate, 3=High, 4=Very high).  These baseline covariates contain demographic and clinical information that is related to baseline disease severity and cytogenetic markers. The primary endpoint in the trial was overall survival time. Our goal is to find the optimal subgroup selection rule that maximizes the size of the selected group while achieving the desired clinically meaningful difference in restricted mean survival time in the survival data. 

    \begin{figure}[!t]
	\centering
	\includegraphics[width=0.5\textwidth]{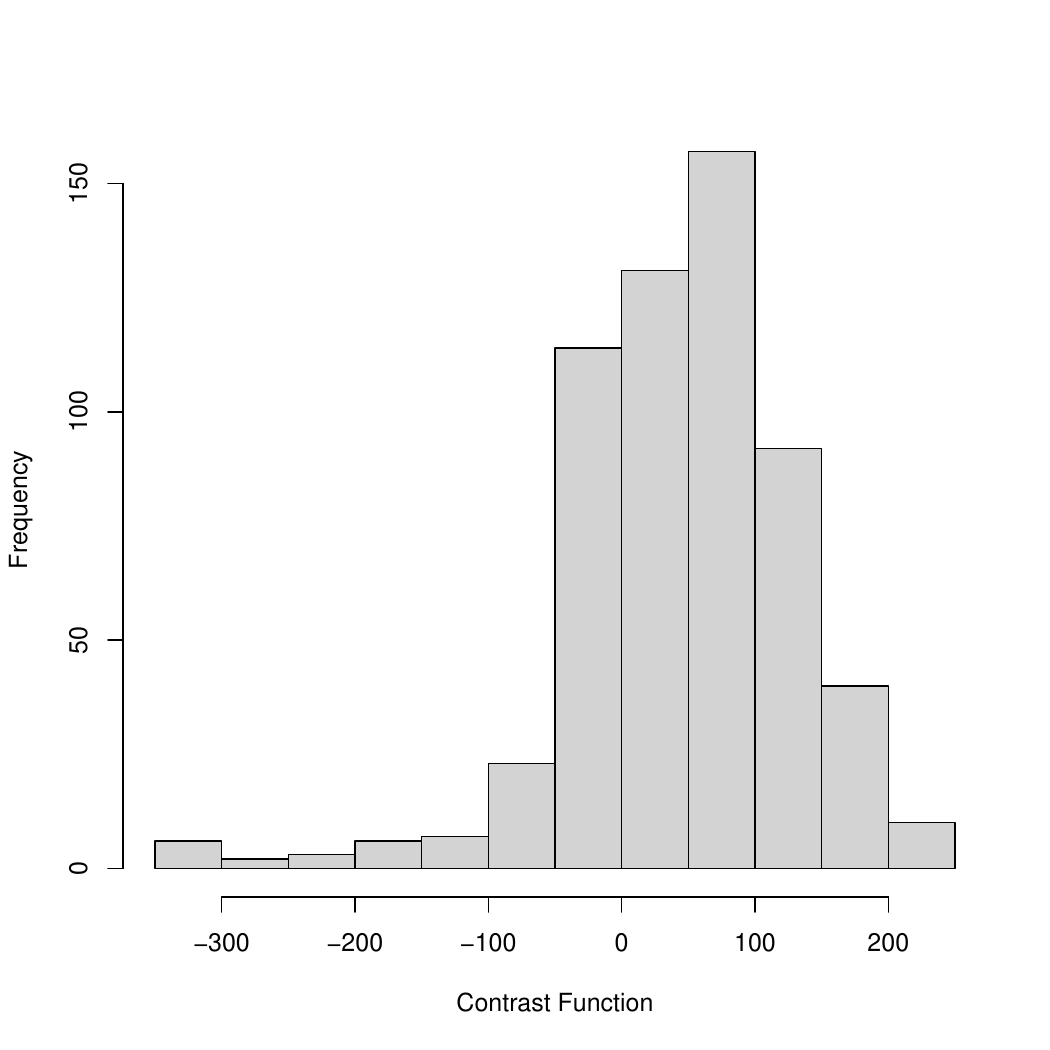} 
	\caption{The density function of the estimated contrast function $\widehat{C}(X)$  for the hematological malignancies data.} \label{fig:lip_den}
\end{figure}

\begin{figure}[!t]
 \centering
\begin{subfigure}{}
	\centering
	\includegraphics[width=0.46\textwidth]{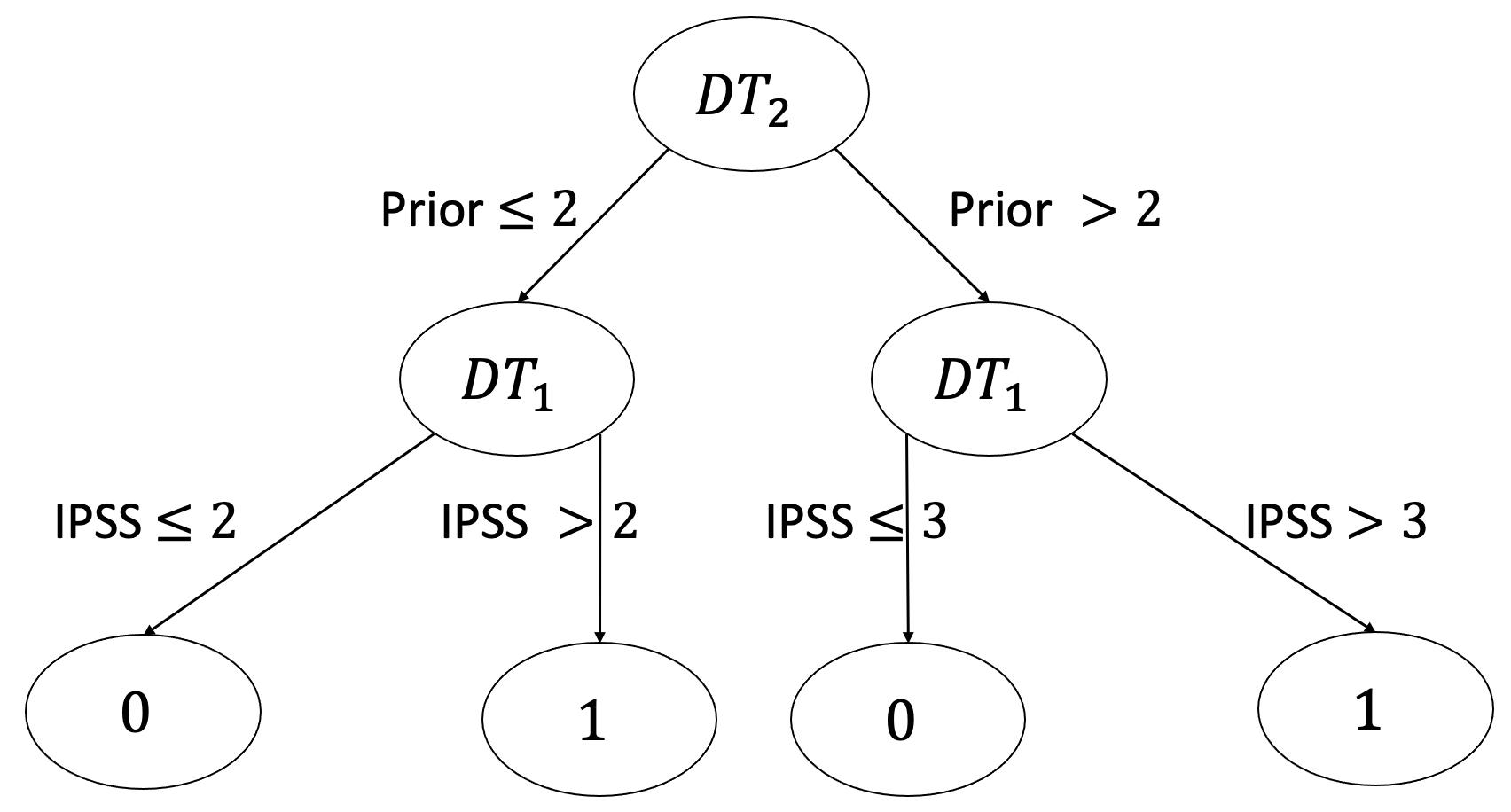}
\end{subfigure}
\begin{subfigure}{}
	\centering
	\includegraphics[width=0.47\textwidth]{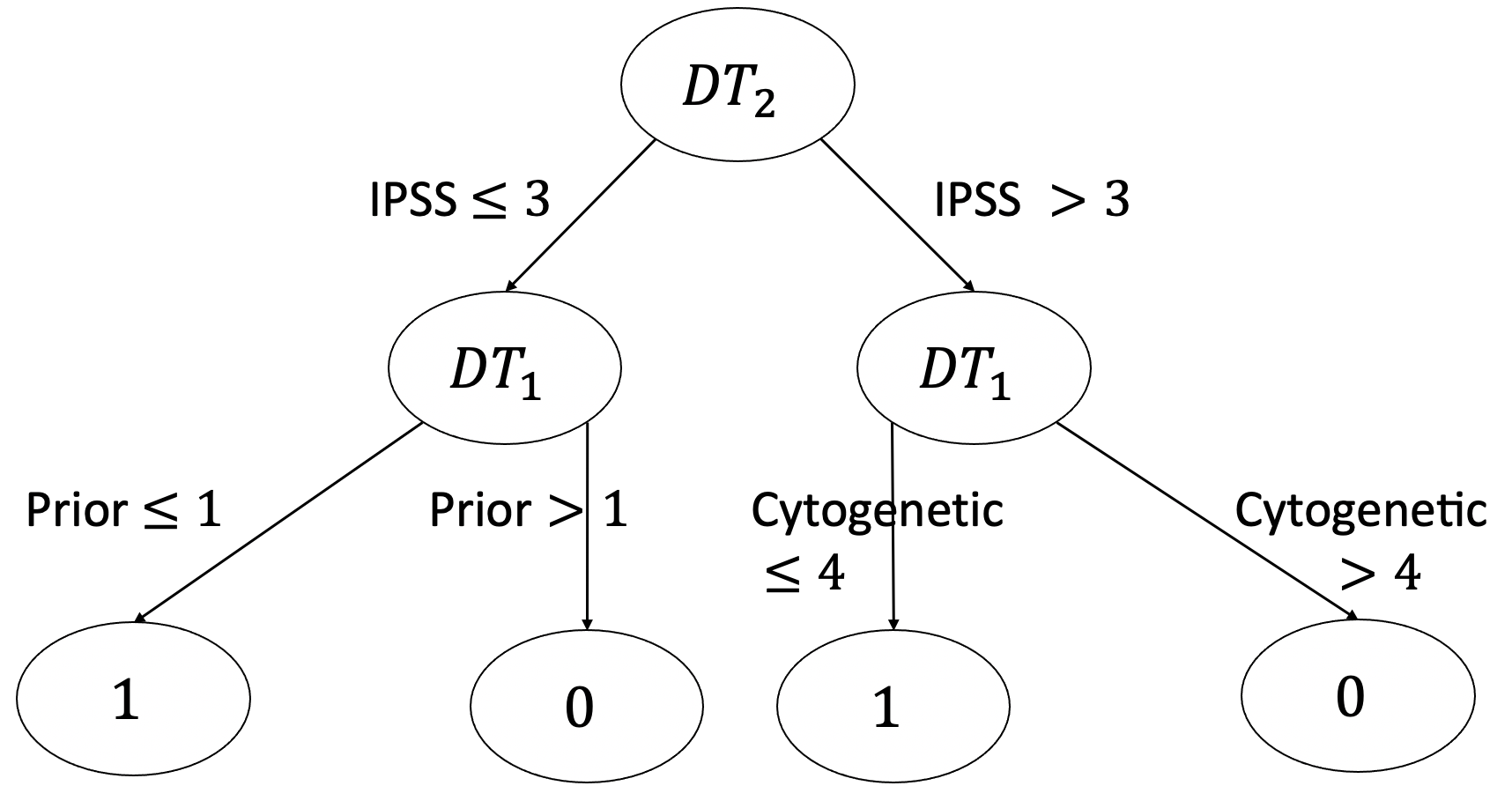}
\end{subfigure} 
	\caption{The estimated optimal subgroup selection tree using CAPITAL  under the hematological malignancies data. Left panel: for $\delta=84$. Right Panel: for $\delta=108$.} 
	 \label{fig:show_dts_lip}
\end{figure}

\begin{table}[!t]
\centering
\caption{Evaluation results of the subgroup optimization using CAPITAL and the subgroup identification (using Virtual Twins \citep{foster2011subgroup}) under the hematological malignancies data.}\label{tab:lip}
\scalebox{0.9}{
 \begin{tabular}{cc|c|c  }
\toprule
 	& & $\delta=84$ & $\delta=108$ \\ 
	\midrule
CAPITAL &$\prob\{\widehat{D}(X)\} $&79.3\% (0.031)&43.2\% (0.057)  \\
 	\cmidrule{2-4} 
 	without penalty &${ATE}(\widehat{D})$&69.5 (5.0)&101.2 (9.8)  \\
		 	\cmidrule{2-4} 
 	&${ATE}(\widehat{D}^c)$&-53.7 (17.7)&1.2 (7.2)\\
		 	\cmidrule{2-4} 
 	&${ATE}(\widehat{D})-{ATE}(\widehat{D}^c)$&123.2 (16.9)&100.0 (9.3)\\
	\cmidrule{2-4} 
	&RPI&87.0\% (0.028)&94.5\% (0.034)   \\
\midrule
	CAPITAL &Penalty $\lambda=$ &2 &2 \\
 	\cmidrule{2-4} 
 	with small penalty  &$\prob\{\widehat{D}(X)\} $&71.7\% (0.061)&33.9\% (0.060)  \\
	\cmidrule{2-4} 
	&${ATE}(\widehat{D})$&74.6 (6.5)&108.4 (9.9)   \\
		 	\cmidrule{2-4} 
 	&${ATE}(\widehat{D}^c)$& -34.1 (15.8)&11.5 (8.7)\\
		 	\cmidrule{2-4} 
 	&${ATE}(\widehat{D})-{ATE}(\widehat{D}^c)$&108.7 (13.2)&96.8 (9.0)\\
	\cmidrule{2-4} 
	&RPI&89.2\% (0.027)&97.2\% (0.034)  \\
\midrule
CAPITAL &Penalty $\lambda=$ &4 & 4 \\
 	\cmidrule{2-4} 
 	with large penalty  &$\prob\{\widehat{D}(X)\} $&51.9\% (0.119)&30.8\% (0.032)\\
 	\cmidrule{2-4} 
 	 &${ATE}(\widehat{D})$&87.2 (13.2)&112.6 (7.0)  \\
		 	\cmidrule{2-4} 
 	&${ATE}(\widehat{D}^c)$&-2.6 (15.9)&13.9 (6.5)\\
		 	\cmidrule{2-4} 
 	&${ATE}(\widehat{D})-{ATE}(\widehat{D}^c)$&89.9 (10.9)&98.7 (8.9)\\
	\cmidrule{2-4} 
	&RPI&92.2\% (0.039)&99.1\% (0.015)   \\ 	
\midrule 
Virtual Twins &$\prob\{\widehat{D}(X)\} $&38.1\% (0.043)&12.9\% (0.117)  \\
 	\cmidrule{2-4} 
      &${ATE}(\widehat{D})$&113.8 (6.2)&151.4 (29.2)  \\
		 	\cmidrule{2-4} 
 	&${ATE}(\widehat{D}^c)$&1.4 (7.2)&29.7 (13.9)\\
		 	\cmidrule{2-4} 
 	&${ATE}(\widehat{D})-{ATE}(\widehat{D}^c)$&112.4 (7.9)&121.7 (21.4)\\
		\cmidrule{2-4} 
	&RPI&99.5\% (0.010)&99.9\% (0.003)   \\
\bottomrule
\end{tabular}}
\end{table}

\textcolor{black}{Based on the estimated contrast function and the estimated difference of mean survival time of 44 days, we consider $\delta = 84$ and $\delta =108$ days, such that the corresponding optimal subgroup sample proportions are approximately around 40\% and 70\%. The density of the estimated contrast function $\widehat{C}(X)$ for the hematological malignancies data is provided in Figure \ref{fig:lip_den}.} 
 We apply the proposed method and the virtual twin method \citep{foster2011subgroup} using the procedure described in Sections \ref{sec:simu_surv} and \ref{sec:real_acgt}. The estimated SSRs under the proposed method are shown in Figure \ref{fig:show_dts_lip}. The evaluation results for the hematological malignancies data are summarized in Table \ref{tab:lip} for varying $\delta $ under the proposed method with $\lambda \in\{0, 2,4\}$ and the virtual twin method.  \textcolor{black}{For Case 1 with ACTG 175 data in Section \ref{sec:real_acgt}, we normalize the outcome by its mean and standard deviation, so the corresponding penalty term is relatively larger compared to Case 2 with the hematological malignancies data in this section, which directly uses the survival time as the outcome.} Our estimated SSRs are shown in Figure \ref{fig:show_dts_lip}, both using the IPSS score and the outcome for the patient's prior therapy as the splitting features in the decision tree. With a desired average treatment effect of $\delta=84$, patients who had a relapse during prior therapy and IPSS larger than 3, as well as those who had no relapse with IPSS larger than 2, are selected into the subgroup with enhanced treatment effect of the experimental treatment plus best supporting care. From Table \ref{tab:lip}, we can also observe that our proposed method has a much better performance compared to the virtual twins method. To be specific, the selected sample proportion under the proposed method is much larger than that under the virtual twins method for all cases, with estimated treatment effect sizes closer to and over the desired clinically meaningful difference in restricted mean survival time as the penalty term $\lambda$ increases. All these findings conform with the results in Section \ref{sec:real_acgt}.

 \section{Discussion}\label{sec:con}
 
 In this paper we proposed a constrained policy tree search method, CAPITAL, to address the subgroup optimization problem. This approach identifies the theoretically optimal subgroup selection rule that maximizes the number of selected patients under the constraint of a pre-specified clinically desired effect. Our proposed method is flexible and easy to implement in practice and has good interpretability. Extensive simulation studies show the improved performance of our proposed method over the popular virtual twins subgroup identification method, with larger selected benefitting subgroup sizes and estimated treatment effect sizes closer to the truth. We further demonstrated the broad usage of our methods for multiple use cases, different trait types, and varying constraint conditions.
 
 \textcolor{black}{The key idea of our proposed algorithm is to transform the constraints defined at the population level into the individual rewards at the patient level. This enables us to identify the patients via policy tree search using rewards as a function of the estimated contrast function $C(\cdot)$. In our numerical studies, we estimate  $C(\cdot)$ based on the random forest method and out-of-bag prediction (Lu et al., 2018 \citep{lu2018estimating}), which is shown to have low bias and variance. We further consider the doubly robust (DR) learner in Step 2 of Algorithm 1 in Kennedy (2020) \citep{kennedy2020optimal} to estimate the conditional average treatment effect with or without additional regression in Equation 2 of Algorithm 1 in Kennedy (2020) \citep{kennedy2020optimal}. The corresponding results are provided in Table \ref{table:tree_dis1} in the appendix based on CAPITAL for Scenario 1 as an illustration, aggregated over 200 replicates. It can be shown from Table \ref{table:tree_dis1} that CAPITAL with the DR-learner plus additional regression achieves comparable results as CAPITAL with the out-of-bag prediction. Yet, without regression, the doubly robust pseudo-outcome in Equation 1 of Algorithm 1 in Kennedy (2020)\citep{kennedy2020optimal} is not consistent at the individual level and thus fails to find the optimal SSR.}
 
There are several possible extensions we may consider in future work. First, we only consider two treatment options in this paper, while in clinical trials it is not uncommon to have more than two treatments available for patients. Thus, a more general method applicable to multiple treatments or even continuous treatment domains is desirable. Second, we only provide the theoretical form of the optimal SSR. It may be of interest to develop the asymptotic performance of the estimated SSR such as the convergence rate.


%
%

   \bibliographystyle{agsm}
\bibliography{mycite}%

\newpage
  \appendix  
  
 \section{Proof of Theorem 1}

The proof of Theorem \ref{thm1} consists of two parts. First, we show the optimal subgroup selection rule is $   \mathbb{I}\{C(x)\geq\eta\}, \forall x \in \mathbb{X}$ where $\eta$ satisfies \eqref{eqn:eta}. Second, we derive the equivalence between \eqref{eqn:ossr1} and \eqref{eqn:ossr2}. Without loss of generality, we focus on the class of SSRs as 
\begin{eqnarray*} 
\Pi\equiv \left[ \mathbb{I}\{C(X)\geq t\}: t\in \mathbb{R}\right].
\end{eqnarray*} 

\textbf{Part One:}  
To show $ \mathbb{I}\{C(x)\geq\eta\}, \forall x \in \mathbb{X}$ is the optimal SSR that solves \eqref{objective}, it is equivalent to show the SSR $ \mathbb{I}\{C(x)\geq\eta\}, \forall x \in \mathbb{X}$ satisfies the constraint in \eqref{objective} and maximizes the size of subgroup. 

First, based on assumptions (A1) and (A2), the average treatment effect under a SSR $ D(X) =  \mathbb{I}\{C(X)\geq t\}$ for a parameter $t$ can be represented by 
\begin{eqnarray*} 
&&\Mean\{Y^*(1)|D(X)=1\}-\Mean\{Y^*(0)|D(X)=1\}= \Mean\{C(X)|D(X)=1\}\\
= &&\Mean[C(X)|\mathbb{I}\{C(X)\geq\eta\}=1] = \Mean\{C(X)| C(X)\geq t \},
\end{eqnarray*}
which is a non-decreasing function of the cut point $t$. Given the definition in \eqref{eqn:eta} that $\Mean\{C(X)|C(X)\geq\eta\}= \delta$, we have $t\in [\eta,+\infty)$ to satisfies the constraint in \eqref{objective}.

Second, the probability of falling into subgroup under the SSR $ D(X) =  \mathbb{I}\{C(X)\geq t\}$ as
\begin{eqnarray*} 
\prob \{D(X)=1\}=  \prob [\mathbb{I}\{C(X)\geq\eta\}=1]=\prob \{ C(X)\geq t \} ,
\end{eqnarray*}
is a non-increasing function of the cut point $t$. 

To maximize the size of subgroup, we need to select the smallest cut point $t$ from its constraint range $t\in [\eta,+\infty)$. Thus, the optimal cut point is $\eta$, which gives the optimal SSR as $\mathbb{I}\{C(x)\geq\eta\}, \forall x \in \mathbb{X}$ as the solution of \eqref{objective}. This completes the proof of  \eqref{eqn:ossr1}. 

\bigskip

\textbf{Part Two:}  
We next focus on proving the optimal SSR in \eqref{eqn:ossr1} is  equivalent to the SSR in \eqref{eqn:ossr2}.  Based on the definition in \eqref{eqn:ossr2}, we have 
\begin{eqnarray*}
D^{1}(x) &&\equiv \mathbb{I}\left(\Mean_{Z\in \mathbb{X}}[ C(Z) \mathbb{I}\{C(Z)\geq C(x)\}] \geq \delta \right)\\
&&=\mathbb{I}\left(\Mean_{Z\in \mathbb{X}}\{C(Z)  | C(Z)\geq C(x) \} \geq \delta \right).
\end{eqnarray*}
Based on the definition in \eqref{eqn:eta} that $\Mean\{C(X)|C(X)\geq\eta\}= \delta$ and the fact that $\Mean\{C(X)|C(X)\geq t\}$ is a non-decreasing function of the cut point $t$, we have the following event holds
\begin{eqnarray*}
\Mean_{Z\in \mathbb{X}}\{C(Z)  | C(Z)\geq C(x) \} \geq \delta,
\end{eqnarray*}
if and only if the following event holds
\begin{eqnarray*}
C(x)   \geq \eta.
\end{eqnarray*}
Thus, it is immediate that the two SSRs in \eqref{eqn:ossr1} and \eqref{eqn:ossr2} are equivalent as the optimal SSR. This completes the proof of  \eqref{eqn:ossr2}.

\begin{sidewaystable}[!thp]
\centering
\caption{Empirical results of subgroup analysis under the estimated optimal SSR by CAPITAL with reward in \eqref{reward2} and the VT-A method.
}\label{table:2}
\scalebox{0.7}{
 \begin{tabular}{cccccc|ccc|ccc}
\toprule
 	Method &&$r=10$&\multicolumn{3}{c|}{Scenario 1} &\multicolumn{3}{c|}{Scenario 2} &\multicolumn{3}{c}{Scenario 3}\\
 	\cmidrule{3-12}
 	&&& $n=200$ & $n=500$&$n=1000$& $n=200$ & $n=500$&$n=1000$& $n=200$ & $n=500$&$n=1000$\\
\toprule
 	CAPITAL&$\delta=0.7$ &Proportion &&$65\%$&&&$67\%$&&&$75\%$&\\ 
 	\cmidrule{2-12}
 	& &$\prob\{\widehat{D}(X)\} $ &0.63(0.16) & 0.63(0.08) & 0.65(0.05) & 0.44(0.24) & 0.52(0.11) & 0.57(0.06) & 0.72(0.15) & 0.75(0.07) & 0.77(0.04) \\
 	\cmidrule{3-12}
 	&&${ATE}(\widehat{D})$& 0.67(0.30) & 0.72(0.17) & 0.70(0.11) & 0.71(0.48) & 0.94(0.20) & 0.85(0.11) & 0.67(0.35) & 0.66(0.17) & 0.60(0.10) \\
	 	\cmidrule{3-12}
 	&&RCD&0.84(0.10) & 0.91(0.05) & 0.93(0.03) & 0.63(0.15) & 0.82(0.08) & 0.87(0.03) & 0.83(0.08) & 0.89(0.03) & 0.91(0.01) \\
	 \cmidrule{3-12}
 	&&RPI&0.78(0.13) & 0.80(0.09) & 0.78(0.06) & 0.74(0.16) & 0.88(0.09) & 0.85(0.07) & 0.67(0.10) & 0.67(0.06) & 0.65(0.04) \\
\cmidrule{2-12} 
 	&$\delta=1.0$ &Proportion &&$50\%$&&&$50\%$&&&$63\%$&\\
 	\cmidrule{2-12} 
	&&$\prob\{\widehat{D}(X)\} $ &0.46(0.16) & 0.48(0.08) & 0.50(0.05) & 0.21(0.18) & 0.32(0.12) & 0.41(0.05) & 0.56(0.16) & 0.60(0.09) & 0.63(0.07) \\
		\cmidrule{3-12}
 	&&${ATE}(\widehat{D})$& 0.91(0.28) & 1.01 (0.15) &0.99(0.10) &0.76(0.66) & 1.32(0.27) & 1.16(0.10) & 1.03(0.39) & 0.99(0.21) & 0.93(0.16) \\
		 	\cmidrule{3-12}
 	&&RCD&0.85(0.11) & 0.92(0.05) & 0.94(0.03) & 0.62(0.12) & 0.79(0.10) & 0.88(0.05) & 0.79(0.08) & 0.85(0.03) & 0.87(0.01) \\
	 \cmidrule{3-12}
 	&&RPI&0.89(0.12) & 0.94(0.06) & 0.95(0.05) & 0.74(0.19) & 0.96(0.03) & 0.97(0.03) & 0.78(0.11) & 0.78(0.07) & 0.76(0.06) \\
\cmidrule{2-12} 
 	&$\delta=1.3$ &Proportion &&$35\%$&&&$37\%$&&&$51\%$&\\
\cmidrule{2-12} 
	&&$\prob\{\widehat{D}(X)\} $ &0.30(0.16) & 0.32(0.11) & 0.34(0.08) & 0.09(0.09) & 0.14(0.10) & 0.25(0.09) & 0.41(0.16) & 0.44(0.09) & 0.48(0.06) \\
 	\cmidrule{3-12}
 	&&${ATE}(\widehat{D})$&1.05(0.35) & 1.27(0.17) & 1.29(0.14) & 0.71(0.76) & 1.57(0.63) & 1.50(0.25) & 1.34(0.42) & 1.40(0.25) & 1.29(0.17) \\
	\cmidrule{3-12}
 	&&RCD&0.81(0.10) & 0.89(0.07) & 0.92(0.04) & 0.67(0.07) & 0.74(0.08) & 0.82(0.06) & 0.77(0.08) & 0.83(0.04) & 0.86(0.02) \\
	 \cmidrule{3-12}
 	&&RPI&0.93(0.13) & 0.99(0.03) & 1.00 (0.01) &0.70(0.21) &0.92(0.14) &0.97(0.02) &0.86(0.10) & 0.90(0.06) &0.88(0.05) \\
\toprule
 	VT-A&$\delta=0.7$ &Proportion&&$65\%$&&&$67\%$&&&$75\%$&\\
 	\cmidrule{2-12} 
 &&$\prob\{\widehat{D}(X)\} $ &0.31(0.12) & 0.34(0.09) & 0.35(0.08) & 0.15(0.10) & 0.19(0.09) & 0.22(0.08) & 0.29(0.10) & 0.30(0.06) & 0.30(0.06)  \\
 	\cmidrule{3-12} 
 	&&${ATE}(\widehat{D})$&  1.11(0.20) & 1.27(0.17) & 1.30(0.15) & 0.85(0.61) & 1.46(0.38) & 1.53(0.32) & 1.76(0.36) & 1.82(0.23) & 1.81(0.21)  \\
	\cmidrule{3-12}
 	&&RCD&0.66(0.12) & 0.69(0.09) & 0.70(0.08) & 0.43(0.08) & 0.51(0.09) & 0.55(0.09) & 0.54(0.10) & 0.55(0.06) & 0.55(0.06)  \\
	 \cmidrule{3-12}
 	&&RPI&0.97(0.06) & 0.99(0.03) & 1.00(0.01) & 0.77(0.17) & 0.95(0.09) & 0.97(0.08) & 0.95(0.07) & 0.98(0.03) & 0.98(0.03)  \\
	\cmidrule{2-12} 
 	&$\delta=1.0$ &Proportion&&$50\%$&&&$50\%$&&&$63\%$&\\
 	\cmidrule{2-12} 
	&&$\prob\{\widehat{D}(X)\} $ &0.21(0.13) & 0.24(0.10) & 0.26(0.07) & 0.07(0.06) & 0.09(0.07) & 0.14(0.07) & 0.23(0.09) & 0.24(0.06) & 0.25(0.05)  \\
 	\cmidrule{3-12}
 	&&${ATE}(\widehat{D})$&  1.19(0.21) & 1.37(0.18) & 1.45(0.13) & 1.01(0.74) & 1.67(0.49) & 1.78(0.38) & 1.94(0.34) & 2.02(0.23) & 2.00(0.18)  \\
	\cmidrule{3-12}
 	&&RCD& 0.70(0.12) & 0.74(0.10) & 0.76(0.07) & 0.54(0.06) & 0.59(0.07) & 0.64(0.07) & 0.60(0.08) & 0.62(0.06) & 0.62(0.05)  \\
	 \cmidrule{3-12}
 	&&RPI&0.98(0.05) & 1.00(0.02) & 1.00(0.00) & 0.81(0.20) & 0.96(0.09) & 0.98(0.08) & 0.97(0.05) & 0.99(0.02) & 0.99(0.01)  \\
	 \cmidrule{2-12} 
 	&$\delta=1.3$ &Proportion&&$35\%$&&&$37\%$&&&$51\%$&\\
 	 \cmidrule{2-12}
	&&$\prob\{\widehat{D}(X)\} $ &0.12(0.11) & 0.11(0.11) & 0.16(0.11) & 0.03(0.04) & 0.03(0.04) & 0.07(0.05) & 0.17(0.09) & 0.18(0.06) & 0.20(0.05)  \\
 	\cmidrule{3-12}
 	&&${ATE}(\widehat{D})$&1.25(0.23) & 1.43(0.18) & 1.50(0.12) & 1.11(0.81) & 1.81(0.61) & 1.98(0.42) & 2.12(0.37) & 2.24(0.23) & 2.19(0.20)  \\
	\cmidrule{3-12}
 	&&RCD& 0.74(0.09) & 0.76(0.11) & 0.81(0.11) & 0.65(0.03) & 0.66(0.04) & 0.69(0.05) & 0.65(0.09) & 0.67(0.06) & 0.69(0.05)  \\
	 \cmidrule{3-12}
 	&&RPI&0.99(0.04) & 1.00(0.01) & 1.00(0.00) & 0.83(0.21) & 0.95(0.13) & 0.98(0.07) & 0.99(0.03) & 1.00(0.01) & 1.00(0.00)  \\
 	\bottomrule
\end{tabular}}
\end{sidewaystable}

\begin{table}[!thp]
\centering
\caption{Empirical results of subgroup analysis under the estimated optimal SSR based on policy tree search by maximizing the $\delta$-quality adjusted value estimator for Scenario 1.
}\label{table:tree_com2}
\scalebox{0.8}{
 \begin{tabular}{ccccc|ccc}
\toprule
 	&Outcome &  \multicolumn{3}{c|}{$Y-\delta$} &\multicolumn{3}{c}{$C(X)-\delta$} \\
 	\cmidrule{1-8}
 	&& $n=200$ & $n=500$&$n=1000$& $n=200$ & $n=500$&$n=1000$ \\
\toprule 
 	\cmidrule{1-8}
 	  Proportion =$65\%$ & $\prob\{\widehat{D}(X)\} $ &0.42(0.04)&0.42(0.03)&0.41(0.02)&0.31(0.11)&0.34(0.06)&0.35(0.04) \\ 
 	\cmidrule{2-8}
 	$\delta=0.7$ &${ATE}(\widehat{D})$& 0.47(0.11)&0.46(0.08)&0.46(0.06) & 1.10(0.27) & 1.26(0.12) & 1.28(0.08) \\
	 	\cmidrule{2-8}
 	&RCD&0.62(0.04)&0.61(0.03)&0.61(0.02)&0.66(0.12)&0.69(0.06)&0.70(0.04) \\
	 \cmidrule{2-8}
 	&RPI&0.70(0.05)&0.70(0.04)&0.71(0.03)&0.95(0.10)&0.99(0.02) & 1.00(0.01) \\
\cmidrule{1-8}
 	  Proportion =$50\%$&$\prob\{\widehat{D}(X)\} $ &0.39(0.04)&0.38(0.03)&0.38(0.02)&0.21(0.12)&0.24(0.08)&0.26(0.05) \\
		\cmidrule{2-8}
 	 $\delta=1.0$ &${ATE}(\widehat{D})$& 0.53(0.13)&0.52(0.10)&0.51(0.07) & 1.11(0.42) & 1.40(0.16) & 1.44(0.09) \\
		 	\cmidrule{2-8}
 	&RCD&0.67(0.04)&0.67(0.03)&0.67(0.03)&0.70(0.12)&0.73(0.08)&0.76(0.05) \\
	 \cmidrule{2-8}
 	&RPI&0.72(0.05)&0.72(0.04)&0.73(0.04)&0.93(0.14)&0.99(0.02 & 1.00(0.00) \\
\cmidrule{1-8}
 	Proportion =$35\%$ &$\prob\{\widehat{D}(X)\} $ &0.34(0.04)&0.34(0.03)&0.34(0.02)&0.12(0.10)&0.12(0.07)&0.15(0.07) \\
 	\cmidrule{2-8}
 	$\delta=1.3$  &${ATE}(\widehat{D})$&0.60(0.14)&0.58(0.10)&0.56(0.07) & 1.07(0.54) & 1.49(0.27) & 1.58(0.11) \\
		\cmidrule{2-8}
 	&RCD&0.69(0.04)&0.69(0.03)&0.68(0.02)&0.75(0.09)&0.76(0.07)&0.80(0.07) \\	 \cmidrule{2-8}
 	&RPI&0.75(0.06)&0.75(0.05)&0.75(0.04)&0.89(0.18)&0.98(0.07 & 1.00(0.01) \\
\bottomrule
\end{tabular}}
\end{table}

 \begin{table}[!thp]
\centering
\caption{Empirical results of optimal subgroup selection tree by CAPITAL plus the doubly robust estimation of $C(\cdot)$ with or without additional regression for Scenario 1.
}\label{table:tree_dis1}
\scalebox{0.8}{
 \begin{tabular}{ccccc|ccc}
\toprule
 	&Method &  \multicolumn{3}{c|}{Without Regression} &\multicolumn{3}{c}{With Regression} \\
 	\cmidrule{1-8}
 	&& $n=200$ & $n=500$&$n=1000$& $n=200$ & $n=500$&$n=1000$ \\
\toprule
 	  Proportion =$65\%$ & $\prob\{\widehat{D}(X)\} $ &0.93(0.05)&0.94(0.04)&0.94(0.04)&0.70(0.15)&0.69(0.09)&0.67(0.05) \\
 	\cmidrule{2-8}
 	$\delta=0.7$ &${ATE}(\widehat{D})$& 0.10(0.08)&0.10(0.08)&0.11(0.07)&0.51(0.27)&0.60(0.18)&0.64(0.11) \\
		 	\cmidrule{2-8}
 	&RCD&0.70(0.05)&0.71(0.04)&0.71(0.04)&0.83(0.08)&0.90(0.06)&0.93(0.03) \\
	 \cmidrule{2-8}
 	&RPI&0.53(0.03)&0.53(0.03)&0.53(0.02)&0.71(0.13)&0.74(0.09)&0.75(0.06) \\
\cmidrule{1-8}
 	 Proportion =$50\%$ &$\prob\{\widehat{D}(X)\} $ &0.86(0.08)&0.81(0.08)&0.78(0.07)&0.52(0.14)&0.53(0.07)&0.52(0.06) \\
		\cmidrule{2-8}
 	$\delta=1.0$&${ATE}(\widehat{D})$& 0.21(0.15)&0.34(0.16)&0.43(0.14)&0.84(0.26)&0.91(0.13)&0.94(0.11) \\
		 	\cmidrule{2-8}
 	&RCD&0.62(0.08)&0.68(0.08)&0.72(0.07)&0.84(0.08)&0.92(0.04)&0.94(0.03) \\
	 \cmidrule{2-8}
 	&RPI&0.57(0.06)&0.62(0.07)&0.65(0.06)&0.86(0.11)&0.90(0.07)&0.93(0.06) \\
\cmidrule{1-8}
 	Proportion =$35\%$ &$\prob\{\widehat{D}(X)\} $ &0.74(0.12)&0.65(0.08)&0.60(0.05)&0.37(0.14)&0.39(0.08)&0.40(0.06) \\
 	\cmidrule{2-8}
 	$\delta=1.3$&${ATE}(\widehat{D})$&0.41(0.22)&0.66(0.16)&0.77(0.10) & 1.06(0.23) & 1.17(0.14) & 1.19(0.10) \\
			\cmidrule{2-8}
 	&RCD&0.58(0.11)&0.70(0.08)&0.74(0.05)&0.83(0.08)&0.90(0.05)&0.92(0.04) \\
	\cmidrule{2-8}
 	&RPI&0.66(0.10)&0.77(0.08)&0.82(0.06)&0.93(0.08)&0.98(0.03)&0.99(0.01) \\
\bottomrule
\end{tabular}}
\end{table}

\end{document}